\documentclass[pdflatex, sn-mathphys-num, iicol]{sn-jnl}

\usepackage{comment}
\usepackage{anyfontsize}

\usepackage{graphicx}%
\usepackage{multirow}%
\usepackage{amsmath,amssymb,amsfonts}%
\usepackage{amsthm}%
\usepackage{mathrsfs}%
\usepackage[title]{appendix}%
\usepackage[table]{xcolor}
\usepackage{textcomp}%
\usepackage{manyfoot}%
\usepackage{booktabs}%
\usepackage{algorithm}%
\usepackage{algorithmicx}%
\usepackage{algpseudocode}%
\usepackage{listings}%

\usepackage{url,hyperref}
\usepackage{color}
\usepackage{booktabs}
\usepackage{tikz}
\usepackage{placeins}
\usepackage{siunitx}
\usepackage[export]{adjustbox}
\usepackage{svg}
\usepackage{epstopdf}
\usepackage{multirow}
\usepackage{makecell}
\usepackage{booktabs,nicematrix}
\usepackage{soul}

\usetikzlibrary{shapes,arrows,positioning,calc}
\setstcolor{brown}

\newcommand{\toremove}[1]{\textcolor{blue}{-text removed-}}

\newcommand{\removed}[1]{}

\newcommand{\edit}[1]{#1} 

\newcommand{\handmotion}[1]{\textcolor{blue}{#1}}
\newcommand{\torsomotion}[1]{\textcolor{violet}{#1}}
\newcommand{\other}[1]{\textcolor{teal}{#1}}
\newcommand{\audio}[1]{\textcolor{brown}{#1}}

\theoremstyle{thmstyleone}%
\theoremstyle{thmstyletwo}%

\theoremstyle{thmstylethree}%

\begin{document}

\title[Dancing with REEM-C]{Dancing with REEM-C: A robot-to-human physical-social communication study}


\author*[1,2]{\fnm{Marie} \sur{Charbonneau}}\email{marie.charbonneau@ucalgary.ca}

\author[3,2]{\fnm{Francisco Javier} \sur{Andrade Chavez}}\email{fandradechavez@tru.ca}

\author[4,2]{\fnm{Katja} \sur{Mombaur}}\email{katja.mombaur@kit.edu}

\affil*[1]{\orgdiv{Mechanical and Manufacturing Engineering}, \orgname{University of Calgary}, \orgaddress{\street{2500 University Drive NW}, \city{Calgary}, \state{AB}, \postcode{T2N 1N4}, \country{Canada}}} 

\affil[2]{\orgdiv{Systems Design}, \orgname{University of Waterloo}, \orgaddress{\street{200 University Ave W}, \city{Waterloo}, \state{ON}, \postcode{N2L 3G1}, \country{Canada}}}

\affil[3]{\orgdiv{Engineering}, \orgname{Thompson Rivers University}, \orgaddress{\street{805 TRU Way}, \city{Kamloops}, \state{BC}, \postcode{V2C 0C8}, \country{Canada}}}

\affil[4]{\orgdiv{Institute for Anthropomatics and Robotics}, \orgname{Karlsruhe Institute of Technology}, \orgaddress{\street{Adenauerring 12}, \postcode{76131}, \city{Karlsruhe}, \country{Germany}}} 


\abstract{Humans often work closely together and relay a wealth of information through physical interaction. Robots, on the other hand, are not yet able to work similarly closely with humans and to effectively convey information when engaging in physical-social human-robot interaction (psHRI). This currently limits the potential of 
human-robot collaboration to solve real-world problems. This paper investigates 
how to establish clear and intuitive robot-to-human communication, while considering human comfort during psHRI. 
We approach this question from the perspective of a leader-follower dancing scenario, in which a full-body humanoid robot leads a human by signaling the next steps through a choice of communication modalities including haptic, visual, and audio signals. This is achieved through the development of a split whole-body control framework combining admittance and impedance control on the upper body, with position control on the lower body for balancing and stepping.
Robot-led psHRI participant experiments allowed us to verify controller performance, as well as to build an understanding of what types of communication work better from the perspective of human partners, particularly in terms of perceived effectiveness and comfort.}




\maketitle





\section{Introduction}\label{sec:introduction}
Close contacts commonly occur between humans, for example when caring for others, navigating crowded environments, or carrying out common tasks. In these instances, physical interaction allows to relay significant amounts of information between individuals. While robots are currently envisioned to help with some of these tasks, their ability to handle close physical contacts with humans is yet limited, especially when it comes to ensuring that interactions are safe, socially acceptable, and convey information clearly.

In this context, a robot that can communicate with humans, and that humans can consider as a partner in an interaction, can be considered a \textit{social robot}~\cite{Dunstan2023CulturalRobotics}. Social robots typically engage in what may be termed social human-robot interaction (\textbf{sHRI}): interactions that involve a cooperative relationship or companionship through connection at an emotional level, dialogue, gaze, or gestures~\cite{Dautenhahn2007HRI}. In other scenarios, robots may engage in physical human-robot interaction (\textbf{pHRI}), involving the exchange of forces, contact, or the perception of the material aspects of the interaction. Combining both types of interaction would lead to physical-social human-robot interaction (\textbf{psHRI}), in which physical contact and presence are included as key aspects of a physically and psychologically engaging partnership that emerges between a human and a robot~\cite{farajtabar2024pathcontactbasedphysicalhumanrobot}.

Historically, most works on human-robot interaction (\textbf{HRI}) or human-robot collaboration (\textbf{HRC}) have mainly involved social robots or cobots that do not engage in direct physical interactions with people during their normal operation. 
Recent years have seen the literature on pHRI and psHRI grow~\cite{farajtabar2024pathcontactbasedphysicalhumanrobot}, for instance with the introduction of strategies to enact leader-follower roles during psHRI, such as in~\cite{Whitsell2017, Kobayashi2021}. As outlined in~\cite{Semeraro2023,Lanini2018}, enabling robots to detect a human partner's intentions is a key research direction towards robots that effectively handle physically interactive tasks. However, little research has yet addressed the opposite problem of enabling robots to clearly communicate intentions to human partners during psHRI. This is not only crucial for effective HRC, but also to ensure the safety and comfort of human partners. Developing the ability of robots to initiate contact and intuitively lead through touch would amplify their effectiveness in assistive applications such as rehabilitation, search and rescue, or alleviating social isolation.

This paper therefore investigates how the actions of a robot affect a human partner's understanding of robot intentions, and feeling of comfort
during psHRI. 
As a test case, we consider a partner dancing task, 
i.e., a self-contained, but highly physically and psychologically engaging activity that requires clear communication between partners. 
Combining a physical component in the exchange of forces and sensorimotor information through direct contact, along with social components in the form of verbal and non-verbal communication and coordination, partnered human-robot dancing is a prime example of psHRI. Studying psHRI through dance is especially relevant, given not only the physical activity and close physical interaction aspects of dance, but also the diversity of social skills that arise through dance, such as ritualized and accurate communication, emotional expression, coordination and cooperation~\cite{Fink2021Evolution, Narikbayeva2025The, Rebrikova2024THE}.
 
\edit{In this direction, we developed a framework that allows a humanoid robot to lead basic dance steps using haptic, visual and audio cues to communicate its intended next step. We then collected feedback from human participants on how they perceived the interaction. The resulting overall contribution of the paper is the introduction of multi-modal whole-body human-robot communication. Given that the underlying framework to achieve communication, as well as the physical and social aspects of the communication are equally important, the contributions of the paper are broken down on three complementary levels\footnote{Note that contributions will be further specified in Sec.~{\ref{sec:related work}}.}
:\\
(i) low-level control: the paper presents a whole-body controller augmenting a balancing and walking controller with upper-body compliance and impedance for multi-modal communication;\\
(ii) high-level control: the paper introduces multi-modal communication strategies for HRI, in the form of leading signals enabling a robot to communicate intention;\\
(iii) knowledge gained: the paper synthesizes lessons learned with the resulting multi-modal communication, and proposes guidelines to facilitate human interpretation when signaling robot intention, while maintaining feelings of comfort. \\
Each of the three sub-contributions can stand on its own: the proposed whole-body controller, communication strategies, and guidelines may each be independently adopted in other HRI scenarios.}

\begin{figure}
    \centering
    \includegraphics[height=0.55\linewidth]{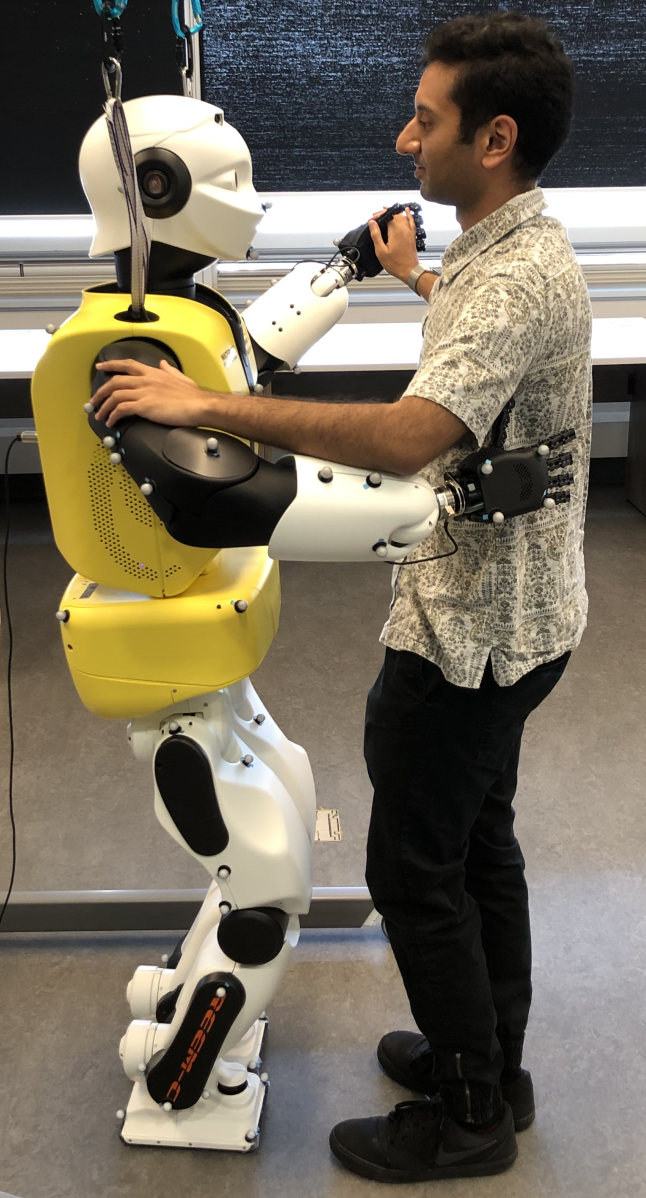}
    \includegraphics[height=0.55\linewidth]{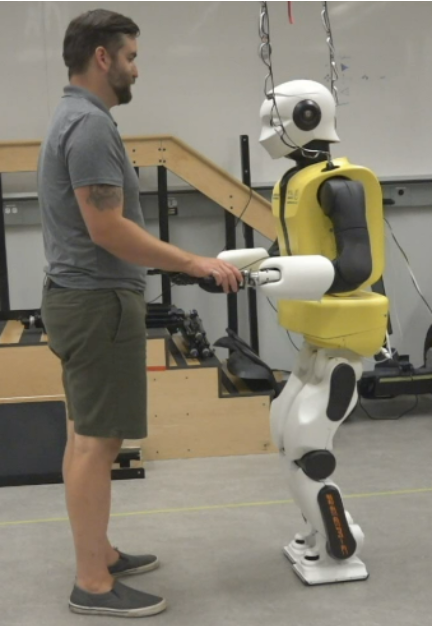}
    \caption{Left: closed position without body contact at the hip or torso, as may be used in dancing tango. Right: open position with only hand contact, as may be used for salsa or swing dancing. This position is the one used in the experiments, where human participants are made to follow basic box steps led by the robot.}
    \label{fig:dance_setup}\vspace{-0.3cm}
\end{figure}

The paper is divided as follows: Sec.~\ref{sec:related work} reviews relevant literature. Sec. \ref{sec:dancing_robot_setup}, \ref{sec:wbc}, \ref{sec:communication} describe the experimental setup used, the whole-body controller implemented and the communication signals employed. 
In Sec. \ref{sec:experiments}, we describe the experimental protocol. Results are then presented in Sec. \ref{sec:results}, followed by discussion and conclusions in Sec. \ref{sec:conclusion}.

\section{Related work}\label{sec:related work}

`Interaction' can be described as events within communication systems~\cite{newcomb_1953_communication}, thus making communication central to HRI. Typically, communication relies on hearing (e.g., via sounds and speech), sight (e.g., via gaze and body motions,
images and lights) and touch (e.g., via haptics or touch screens)~\cite{bonarini2020communication}. 

Humans are known to adapt how they communicate with a robot based on mental models of the robot~\cite{Kiesler2005commonGround}. At times, individuals may build incorrect mental models,
which can be an obstacle to HRI~\cite{Koening2010communication}. Among approaches proposed to improve human mental models of robots, are the use of auditory cues that inform about the robot's internal state~\cite{Koening2010communication}, and to adapt robot design based on models of communication in HRI~\cite{Frijns2023communication}. Furthermore, it has been shown that when it comes to user satisfaction, communicativeness and responsiveness in a robot matter more than task performance~\cite{Hamacher2016ExpressiveCommunication}.

For more than twenty years now, researchers have been exploring how to establish communication between humans and robots, predominantly through speech and gestures, and primarily human-to-robot communication for robot control~\cite{kanda2003body, Berg2020HRIinterfaces} based on intent detection~\cite{Losey2018pHRCcommunication}. Robot-to-human communication, instead, has been less investigated.
Robot motions and flashing coloured lights have been used to relay information with non-humanoid underwater robots in~\cite{Fulton2019underwaterCommunication}. Speech has been used to query a human operator for information in~\cite{Kaupp2010querying}; verbal commands have been implemented to prompt actions from a human operator in~\cite{Nikolaidis2018communication}; verbal, auditory, visual and haptic cues have been investigated to provide information relative to the robot's internal state or the environment in~\cite{Losey2018pHRCcommunication,Nikolaidis2018communication}; and haptic feedback has been used to guide human movement, for instance when training new motor skills~\cite{Losey2018pHRCcommunication}. 

The reviews in~\cite{Berg2020HRIinterfaces, Losey2018pHRCcommunication} in particular highlight that combining multiple modalities can improve the accuracy of human command recognition and complex task performance in a HRC system.
Nonetheless, to this date, haptics and direct physical interaction remain a promising, yet mostly unexplored communication modality in HRI~\cite{Urakami2023communication}.

Most of the studies cited above assume a human operator as a leader,
but different leader-follower schemes have been used to arbitrate HRC.
Assigning the leading role to a robot resulted in reduced human effort in HRC, and role switching has been used to resolve conflict between human and robot intentions~\cite{Losey2018pHRCcommunication, Li2015lead-follow, Leskovar2021lead-follow}. In~\cite{Evrard2009lead-follow},
a robot is by default following,
but switches to
leading to prevent self-collision.

Role switching behaviours in HRC are studied in~\cite{Whitsell2017,vanZoelen2020lead-follow}. The latter found that while individuals adopt different leading/following strategies (a finding corroborated by~\cite{Holmes2022lead-follow}), one's appreciation of the collaboration 
increases with the number of trials done with the robot, although those who tend to follow more have an overall higher appreciation. 
Similarly, people 
appear to prefer collaborating with a leading robot over one that follows~\cite{Leskovar2021lead-follow}, where predictable robot motions help reduce total workload and increase task performance. Additionally, when a robot is leading, haptic feedback has been shown to help with coordination~\cite{Liu2023haptic-lead-follow}. 
As highlighted in~\cite{Sawers2017dancerForces, Niewiadomski2015nonverbalLeadership}, sudden movements and small interaction forces are commonly used  
in human-human interactions 
such as dancing 
to communicate movement and timing, while a follower is also likely to seek visual cues by looking at the leader.

These latter findings are however not always integrated 
into dancing robots:
\cite{Wang2012robotdancer} used admittance control to minimize interaction forces when a wheeled robot base follows a leading human partner.
Similarly, in~\cite{Chen2015dancingRobot}, the velocity of a mobile base is made to follow dance steps through admittance control based on measured interaction force.
In other works, the focus is instead on adapting a robot's base displacement to a human partner's stride length, based on measured interaction forces: for a following~\cite{Takeda2007,Takeda2007IROS}, or a leading robot~\cite{Holldampf2010HapticDance}. Distinctly, touch semantics for pHRI are proposed in~\cite{Wong2022touchsemantics}, potentially allowing a human to lead a dance with a humanoid.

Literature in which robots lead a dance is still sparse. In~\cite{gentry2005dancingThesis}, a haptic device is used to demonstrate that physical interaction can be sufficient to lead a dance. In~\cite{Granados2017}, a leading dancing robot is used to teach skills to a human learner, based on a closed dancing position (see Fig.~\ref{fig:dance_setup}) with body contact at the hip. In this case, the mobile robot is made to: (i) maintain stiff arms, (ii) use audio signals to count each step, (iii) move its torso up and down with the steps, and (iv) reduce the damping gain of the mobile base controller, as well as interaction force at the hip as learning progresses.

Notably, few psHRI studies are done with legged humanoid robots, and if so, tasks are kept simple. In~\cite{Kobayashi2021}, the focus is on whole-body control and handling conflicts between robot objectives and human objectives through pHRI communication: given a predefined box step dance pattern and force sensors covering its body, the robot is made to take larger strides when it is pushed with larger forces by a human partner. Other humanoids are gradually made to move their base during pHRI such as in~\cite{cheng2024expressiveUnitree,MirokaDancing2025}, but given the lack of footwork and details on the interaction, considering it partner dancing may be debatable.

In all works cited above, authors 
had a predefined idea of interactive robot behaviour,
and did not investigate whether this is what works best for humans. For instance, it seems that the prevalent attitude is that a robot follower is passively compliant, and a robot leader is stiff. However, this principle does not generally translate from human-human interactions, such as dancing. For instance, as emphasized in~\cite{gentry2005dancingThesis}, physical interaction in dancing goes beyond 
the 
communication of movements and coordination: it is also crucial for exchanging energy and influencing the dancers' dynamics. The objective of this paper therefore is to explore what robot behaviours are preferable when a robot is made to lead a human partner to ensure their comfort and safety, including how energy may be deliberately exchanged and control shared in psHRI.

Towards evaluating human preferences in a psHRI scenario, the literature contains a growing collection of validated questionnaires measuring aspects of human-robot interaction. Some capture individuals' general attitudes towards robotics, such as the Negative Attitudes towards Robots Scale (NARS)~\cite{NARS_2006} which includes the statement ``I would feel very nervous just standing in front of a robot'', and the General Attitudes Towards Robots Scale (GAToRS)~\cite{koverola2022_GAToRS} which includes the statement ``I don’t want a robot to touch me''. 
However, these questionnaires do not appear to consider the possibility of psHRI and humans communicating with a robot through touch. 

Other scales look at robot attributes relating to appearance, anthropomorphism, animacy, likeability, perceived intelligence, or perceived safety, part of which are directly relevant to psHRI. For instance, the Godspeed questionnaire series~\cite{bartneck2009godspeed} invite respondents to rate their impression of a robot on scales including machinelike/humanlike, moving rigidly/elegantly, apathetic/responsive. In the Robotic Social Attribute Scale (RoSAS)~\cite{Carpinella2017RoSAS}, respondents evaluate how closely they associate robots with words related to competence (e.g., responsive, interactive), warmth, and discomfort (e.g., awkward, aggressive).

The trust measurement scale for industrial human-robot collaboration~\cite{charalambous2016trust_in_HRC} includes relevant statements relating to comfort (``The way the robot moved made me uncomfortable'', ``I was comfortable the robot would not hurt me'', ``The size of the robot did not intimidate me'', ``I felt safe interacting with the robot''), alongside statements relating to the accomplishment of an industrial task. 
The questionnaire for the evaluation of physical assistive devices (QUEAD)~\cite{schmidtler2017QUEAD} measures usability and acceptance aspects in pHRI with a focus on robot control modes and the accomplishment of tasks, with statements such as ``I feel comfortable using this control mode''. 
The Human-Robot Interaction Stress Scale~\cite{babamiri2024HRISS}, while published after the experiments presented in this paper took place, includes statements relating to human-robot communication (e.g., ``When working with the robot, I must pay close attention in order to obtain the necessary information'', ``There are times when I am unable to interpret the information obtained when working with the robot'') alongside statements targeted towards occupational stress when operating a robot. 

Despite these advances, we did not find a questionnaire adapted to evaluating comfort or quality of communication within a psHRI scenario. 

\section{REEM-C humanoid robot and psHRI abilities}\label{sec:dancing_robot_setup}

To understand how humans experience being led by a robot during psHRI, a few essential building blocks first need to be selected and assembled: what robot will be used (since hardware affects what interactions are safe and feasible), what physical interaction will be studied, and how it will be implemented.

\subsection{The robot}\label{subsec:robot}

The robot used is the REEM-C legged humanoid robot from PAL Robotics~\cite{noauthor_reem-c_nodate}. At 1.65 \si{m} tall and 76 \si{kg} heavy, it comes with 28 actuated joints for whole-body control: 7 in each arm, 6 in each leg, and 2 in the torso. Given the lack of joint torque sensors, each joint is position-controlled. The robot is equipped with a 6-axis force/torque (\textbf{F/T}) sensor in each wrist and ankle, allowing to measure physical interactions with the environment at the hands and feet of the robot.

By default, the REEM-C comes with a walking controller based on~\cite{kajita_biped_2003}, including a stabilizer inspired by~\cite{kajita_biped_2010}. Using this controller, the robot can be made to execute one step per second (inclusive of single and double support phases), given what foot to move and by how much. Step length can be speficied in terms of forward/backward and left/right directions.

The face of the REEM-C is made of a solid shell: its `eyes' are two black rounds; its mouth a white rectangle. Thus, it cannot show facial expressions. The robot also comes with a speech synthesizer, allowing the robot to `speak' given input text, although without the ability to modulate tone. 
\edit{When commands are sent to the robot to move a body part, step, or synthesize speech, the robot sequentially executes each action after the other; the timing between speech and robot movement may not be explicitly set by the user.}

\subsection{The interaction}\label{subsec:interaction}

Given current hardware capabilities, the robot can support interactions that include (i) forces exchanged with a human through the hands, and (ii) simple
footwork at a frequency of 60 steps per minute.
A basic box step pattern at a pace similar to that of Argentine tango and danced in an open position, as described in the next paragraphs, would fully exploit these capabilities. 

In the open position, shown in Fig.~\ref{fig:dance_setup}, partners stand apart facing each other, and the follower's hands are placed palm down on the leader's palm up hands to establish connection.

\begin{figure}[t!]
    \centering
    \includegraphics[width=\linewidth]{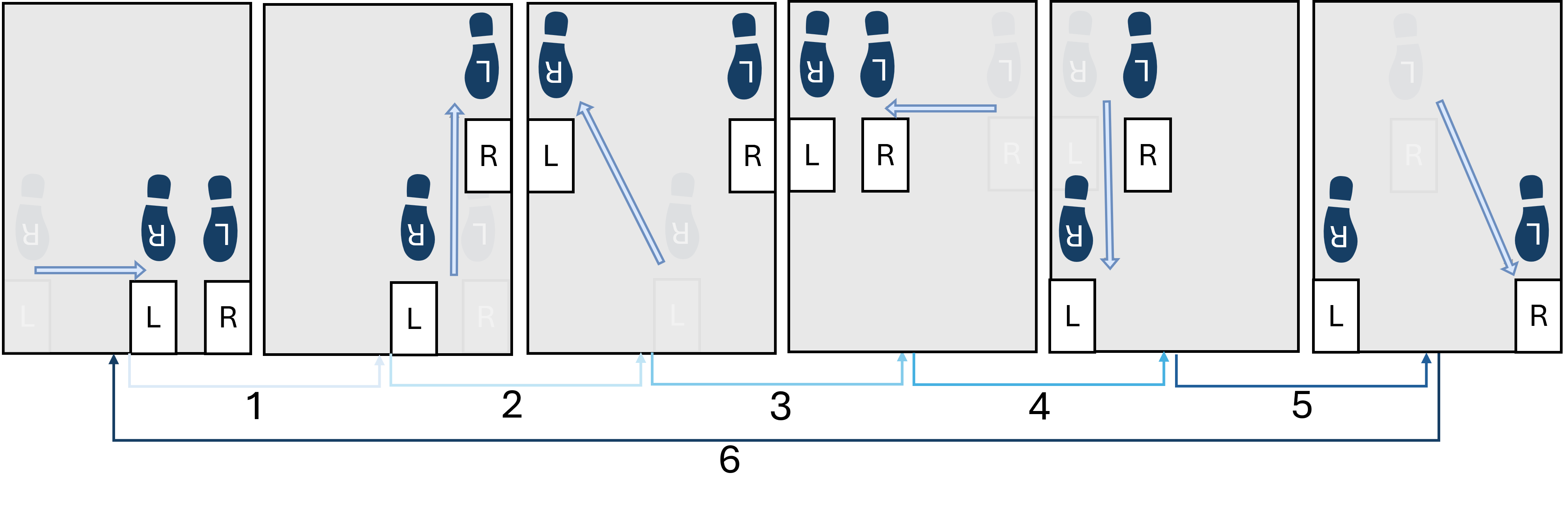} 
    \vspace{-0.5cm}
    \caption{Step sequence for the interaction. White boxes denote robot (leader) feet; footstep shapes denote the human's (follower).}
    \label{fig:step_sequence}
\end{figure}

The box step footwork is shown in Fig.~\ref{fig:step_sequence} for both leader and follower. It consists of a sequence of 6 steps making a square shape on the floor, with forward, backward and diagonal foot displacements.

As the leader, the robot will be executing the leader's steps and setting the timing of the dance. 
The role of the human follower is to interpret the dance by coordinating the timing and trajectory of their steps in response to cues perceived from the robot.

While musical cues can influence the coordination of dancers, physical communication between the leader and the follower is more important for the follower to select what
step to take~\cite{Gentry2004Musicality}. When learning a dance, dancers typically start without music and by repeating a chunk of movements (such as a sequence of steps), to focus on motion and coordination between partners, while allowing muscle memory to form\footnote{As per the first author's experience learning over 10 dance styles, from early childhood to the present day.}.
Thus, we choose to conduct experiments in this way to facilitate participants' focusing their attention on the experience of being led through psHRI. This will help support the objective of investigating how different robot communication behaviours affect human interpretation and comfort during psHRI.
Given that choice, participants could technically anticipate the next step. However, this in no way restricts participants' ability to perceive cues from the robot and offer their insights on them, to make errors, and to develop their own motion and timing.

\section{Whole-body control for psHRI}\label{sec:wbc}

\begin{figure*}[t!]
    \centering
    \includegraphics[width=\linewidth]{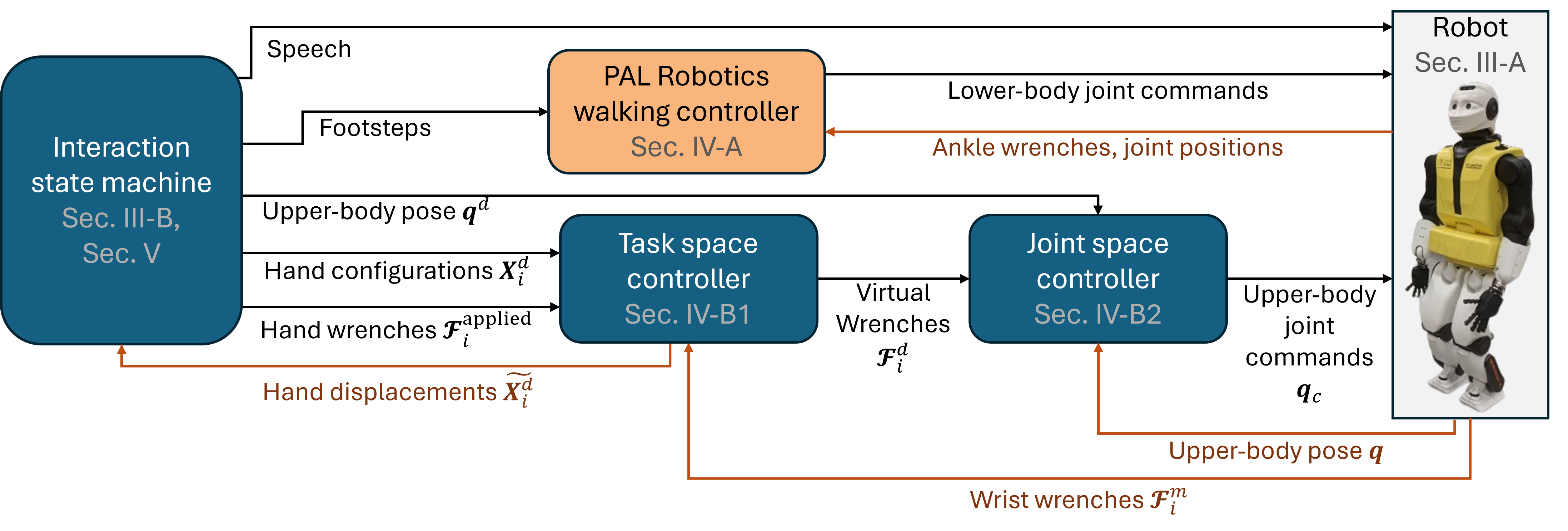}
    \caption{Whole-body control architecture for psHRI} \vspace{-0.3cm}
    \label{fig:control}
\end{figure*}

While safety in psHRI is typically ensured with passive compliant control~\cite{farajtabar2024pathcontactbasedphysicalhumanrobot}, as introduced in Sec.~\ref{sec:related work}, robots are generally made to lead with stiff control. Thus, a compromise is needed to enable safe and effective communication through haptics, while stepping and maintaining balance. We accomplish this with a whole-body controller combining admittance and impedance control on the upper-body, and position control on the lower-body, as shown in Fig.~\ref{fig:control} and detailed below. 

\subsection{Lower-body control} \label{subsec:Lower-body-control}

The default walking controller from PAL Robotics described in Sec.~\ref{subsec:robot} is applied to the leg joints, allowing to maintain balance and to move the feet of the robot. This controller only affects the leg joint motions, but the stabilizer is designed to compensate for upper body motions, including those generated by the controller described below.

\subsection{Upper-body control} \label{subsec:Upper-body-control}

For the arms and torso joints, a controller is defined to enable the following options: (i) tracking desired hand trajectories in task space, (ii) tracking desired upper-body joint trajectories in configuration space, (iii) passive compliance to forces applied at the hands of the robot, and (iv) actively applying desired interaction forces at the hands. These goals are accomplished with a hierarchical controller, as part of which a task space controller feeds into a joint space controller, in order to compute the joint position commands that are sent to the robot.

\subsubsection{Task space controller}\label{subsubsec:task space control}

The idea is to compute a virtual wrench $\boldsymbol{\mathcal{F}}_i^d~\in~\mathbb{R}^6$ including the desired 3D forces and moments to be applied at each hand $i$ of the robot. This wrench is defined in terms of the base frame, situated in the centre of the base link of the robot (i.e., the ``pelvis'' link, serially connecting to the torso and the legs), with the $x$-axis pointing forward, the $z$-axis upward, and the $y$-axis forming a right-handed frame.

For passive compliance, a simplified admittance control law relying on the wrench $\boldsymbol{\mathcal{F}}_i^m$ measured by the F/T sensor attached to hand $i$ is defined as:
\begin{equation}
    \boldsymbol{\mathcal{F}}_i^\text{admittance} = \boldsymbol{G}_T \boldsymbol{\mathcal{F}}_i^m
\end{equation}
where $\boldsymbol{G}_T$ is a diagonal gain matrix modulating admittance. 

To move hand $i$ toward a desired configuration $\boldsymbol{X}_i^d \in \mathbb{R}^6$ with respect to the base frame, an impedance control law is applied:
\begin{equation}
    \boldsymbol{\mathcal{F}}_i^\text{impedance} = - \boldsymbol{K_{P_T}} \boldsymbol{\widetilde{X}}_i - \boldsymbol{K_{D_T}} \boldsymbol{\widetilde{\dot{X}}}_i
\end{equation}
with $\boldsymbol{K_{P_T}}$, $\boldsymbol{K_{D_T}}$ diagonal proportional and derivative gain matrices, and $\boldsymbol{\widetilde{X}}_i$, $\boldsymbol{\widetilde{\dot{X}}}_i$ the error between the current and desired hand configuration and velocity. 

To actively apply a specified wrench at a hand, a term $\boldsymbol{\mathcal{F}}_i^\text{applied}$ may directly be used. 

From there, the three virtual wrenches are added together to obtain $\boldsymbol{\mathcal{F}}_i^d$, the output of the task space controller:
\begin{equation}
    \boldsymbol{\mathcal{F}}_i^d =  \boldsymbol{\mathcal{F}}_i^\text{admittance} + \boldsymbol{\mathcal{F}}_i^\text{impedance} + \boldsymbol{\mathcal{F}}_i^\text{applied}
\end{equation}

In practice, because the admittance and impedance components by definition act against each other, 
we impose\footnote{Sharp discontinuities in the controller are avoided by gradually updating values over a desired transition time $t_t$.} 
$\boldsymbol{K_{P_T}} =\boldsymbol{K_{D_T}} = \boldsymbol{0}_{6\times6}$ when $\boldsymbol{\mathcal{F}}_i^m \geq \boldsymbol{\epsilon}_\mathcal{F}$, for a given set of force and moment thresholds $\boldsymbol{\epsilon}_\mathcal{F}$. 
Similarly, to prevent $\boldsymbol{\mathcal{F}}_i^\text{applied}$ from interfering with the impedance controller, elements of $\boldsymbol{K_{P_T}}$ and $\boldsymbol{K_{D_T}}$ corresponding to the axes along which a non-zero force or moment is applied are set$^3$ to zero. For example, if hand $i$ applies a diagonal force along the $x$- and $y$-axes such that $\boldsymbol{\mathcal{F}}_i^\text{applied} = \begin{bmatrix} f_x & f_y & 0 & 0 & 0 &0  \end{bmatrix}^\top$, then $\boldsymbol{K_{P_T}} = \begin{bmatrix}
    0 & 0 & k_{f_z} & k_{m_x} & k_{m_y} & k_{m_z}
\end{bmatrix}^\top$.

\subsubsection{Joint space controller}
Joint commands are computed to realize the virtual wrench output from the task space controller, as well as to move joints towards desired positions.

First, $\boldsymbol{\mathcal{F}}_i^d$ is projected onto the $n$ revolute joints of the torso and arm, to get the virtual joint torques $\boldsymbol{\tau^\text{admittance}}~\in~\mathbb{R}^n$ required to produce the virtual wrenches obtained for each hand $i$, with:
\begin{equation}
    \boldsymbol{\tau^\text{admittance}} = \frac{1}{n_h} \sum_{i=1}^{n_h} \boldsymbol{J}_i^\top \boldsymbol{\mathcal{F}}_i^d
\end{equation}
where $n_h$ is the number of hands in contact.
$\boldsymbol{J}_i^\top~\in~\mathbb{R}^{n \times 6}$ is the transpose of the Jacobian matrix that maps upper-body joint velocities to hand $i$ twist (the 6D vector of linear and angular velocity). 
Normalizing by $n_h$ ensures that wrenches coordinated between hands are not applied multiple times.

For the latter objective, an impedance control law is used:
\begin{equation}
    \boldsymbol{\tau}^\text{impedance} = 
        -\boldsymbol{K_{P_J}} \boldsymbol{\tilde{q}}
        - \boldsymbol{K_{D_J}} \boldsymbol{\tilde{\dot{q}}}
\end{equation}
with $\boldsymbol{K_{P_J}}$, $\boldsymbol{K_{D_J}}$ diagonal proportional and derivative gain matrices; $\boldsymbol{\tilde{q}}$, $\boldsymbol{\tilde{\dot{q}}}$ the error between the current and desired joint positions and velocities.
Similarly as in~\cite{Shingarey2020tbvc}, we then apply:

\begin{equation}
    \boldsymbol{\dot{q}}_c = \boldsymbol{G}_J^a \boldsymbol{\tau^\text{admittance}} + \boldsymbol{G}_J^i \boldsymbol{\tau}^\text{impedance}
\end{equation}
such that virtual torque feedback is used to adjust the input of the underlying joint velocity controller. $\boldsymbol{G}_J^a$, $\boldsymbol{G}_J^i$ are diagonal gain matrices modulating admittance and impedance. We chose a constant $\boldsymbol{G}_J^a = \mathbb{I}$. We set $\boldsymbol{G}_J^i = \mathbb{I}$ when $\boldsymbol{\mathcal{F}}_i^m < \boldsymbol{\epsilon}_\mathcal{F}$,  (i.e., $\boldsymbol{G}_J^i$ is set to the identity matrix when a negligible wrench is applied to either hand), otherwise $\boldsymbol{G}_J^i$ is decreased$^3$ to predefined minimum values. Doing so prioritizes compliance and safety, while gradually resisting applied forces the further away joints are moved from desired positions.

The joint commands $\boldsymbol{q}_c$ are then obtained by integrating $\boldsymbol{\dot{q}}_c$. 
In the process, $\boldsymbol{\dot{q}}_c$ and $\boldsymbol{q}_c$ are clamped to respect joint velocity and position limits.

\section{Communication for a leading robot} \label{sec:communication}

For the robot to lead the interaction described in Sec.~\ref{subsec:interaction}, robot-to-human communication strategies need to be defined. Given the current abilities of the REEM-C, it can be made to communicate through forces applied at the hands, body motions, or speech. Each would be perceived by a human partner either as a haptic, visual or audio cue. To understand what modalities work best during psHRI, we defined, \edit{in informal consultation with dancers,} a set of `leading signal' candidates to cue a human partner as to what foot to step with, and in which direction. As detailed in Table~\ref{tab:lead-signals}, leading signals include haptic cues through the hands, visual cues through upper- and lower-body motion, and audio cues through speech.

\begin{table*}[hbt!]
    \centering
    \caption{Types of leading signals used (separately or in combinations with each other). The top 6 are leading signals deliberately defined for this study, while the bottom one indicates visual cues that are inherent to the dancing scenario.}

    \rowcolors{2}{white}{gray!20}
    \begin{tabular}{ p{3cm} p{1cm} p{10.75cm} }
    \hline
    \textbf{Title}         & \textbf{Type}    & \textbf{Description}   \\ 
    \hline

    \raggedright Hand wrench (\textbf{HW}) & Haptic & 
    Apply virtual wrench $\boldsymbol{\mathcal{F}^\text{applied}_i}$ at the hand $i$ corresponding to the stepping foot, in the direction and for the duration of the step. Given the steps of Fig.~\ref{fig:step_sequence}, this results in wrenches composed of horizontal forces:~
    \makecell{
    forward/backward step: $\boldsymbol{\mathcal{F}^\text{applied}_i}~=~[
        \pm f^d ~ 0 ~ 0 ~ 0 ~ 0 ~ 0 ]^\top$, \\
    left/right step: $\boldsymbol{\mathcal{F}^\text{applied}_i}~=~[
        0 ~ \pm f^d ~ 0 ~ 0 ~ 0 ~ 0 ]^\top$, \\
    diagonal step: $\boldsymbol{\mathcal{F}^\text{applied}_i}~=~[
        \pm f^d ~ \pm f^d ~ 0 ~ 0 ~ 0 ~ 0]^\top$.}
    \\

    \raggedright Hand displacement (\textbf{HD}) & Haptic & Apply a displacement of $\Delta\boldsymbol{X}_i^d$ at the hand $i$ corresponding to the stepping foot, in the direction and for the duration of the step. The steps of Fig.~\ref{fig:step_sequence} result in horizontal displacements:~
    \makecell{
    forward/backward step: $\Delta\boldsymbol{X}_i^d~=~[
        \pm \delta^d ~ 0 ~ 0 ~ 0 ~ 0 ~ 0 ]^\top$, \\
    left/right step: $\Delta\boldsymbol{X}_i^d~=~[
        0 ~ \pm \delta^d ~ 0 ~ 0 ~ 0 ~ 0 ]^\top$, \\
    diagonal step: $\Delta\boldsymbol{X}_i^d~=~[
        \pm\delta^d ~ \pm\delta^d ~ 0 ~ 0 ~ 0 ~ 0]^\top$.}
    \\

    \raggedright Torso rotation (\textbf{TR})  & Visual  & Apply a rotational displacement $\Delta q_{\text{torso}}$ at the torso yaw joint, such that the shoulder corresponding to the stepping foot moves in the direction and for the duration of the step, but keeping hands in the same configuration $\boldsymbol{X}_i^d$ with respect to the base. 
    The head rotates in a block with the torso. \\

    \raggedright Step count (\textbf{SC})     & Audio     & Make the robot verbally count each step from 1 to 6, before taking the corresponding step. \\ 

    \raggedright Step description (\textbf{SD})       & Audio     & Make the robot verbally describe what the human partner should be doing (e.g., ``Step forward'', ``Step left''). \\

    \raggedright No signal (\textbf{NS})  & N/A  & No specific cue deliberately used to communicate actions, i.e., $\boldsymbol{\mathcal{F}^\text{applied}_i}~=~\Delta\boldsymbol{X}_i^d = \boldsymbol{0}_6$ and $\Delta q_{\text{torso}} = 0 $.\\

    \hline
    \raggedright Foot and base displacement & Visual & The stepping foot moves in order to accomplish a given step. The base of the body follows, with the upper-body moving in a block (i.e., if the base moves in a given direction by a certain distance with respect to an inertial frame, so do the hands). \textbf{Invariably occurring at each step, it is considered a constant}.\\

    \hline
    \end{tabular}
    \label{tab:lead-signals}
\end{table*}

Haptic and visual leading signals were defined considering that muscle tone and an upright upper-body shape are essential components of a frame that makes leading and following possible. The use of either displacement or force at the hand will affect the perceived ``muscle tone'' of the robot. Towards maintaining a coherent upper-body shape, the controller defined in Sec.~\ref{subsec:Upper-body-control} allows the torso to rotate in the direction in which hands are translated. Linear motion of the torso also naturally follows from the translation of the base  when stepping the feet. 

Note that while the robot controller can generate wrenches and displacements at the hands in any direction (subject to the actuation limits of the robot), the dance steps as introduced in Fig.~\ref{fig:step_sequence} only require horizontal forces and translations. \edit{The chosen leading signals were limited in this respect. They were intentionally kept simple in this study, to ensure participant safety and to remain legible to those who have a limited dance background. Other options could be explored in the future, with caution.
For example, when well executed,} rotating the hand around the upper body would indicate a turn to the follower, but perhaps rotating the hand in place would simply twist the follower's wrist. 

One could also have considered creating additional left/right or forward/backward motion of the torso as a leading signal by translating or leaning from the base. However, (i) this would potentially be questionable form in leading the selected steps, (ii) torso translation with respect to the base is infeasible with the articulations of the REEM-C, and (iii) leaning would reasonably risk communicating a lack of balance from the robot, rather than what step to take.

Communication from the human partner to the robot is somewhat limited with our implementation. However, the admittance control allows one to move the hands of the robot around with little resistance. For instance, if someone is uncomfortable or not following the robot, they may push the hands of the robot away. A hand deflection $\boldsymbol{\widetilde{X}}_i$ larger than a threshold 
is used as a signal for the robot to stop moving.


\section{Participant experiments} \label{sec:experiments}
Once the controller was verified to perform as expected, participant experiments were set up to understand how individuals perceive the different leading signals in the physical interaction introduced in Sec.~\ref{subsec:interaction}.

\subsection{Participant recruitment}\label{sec:participants}
We recruited $n_p=22$ participants through word of mouth within the University of Waterloo and local dance communities. What we told prospective participants was that the study involves holding hands with a robot while dancing. To be included in the study, participants had to be
in good health such that it would not affect their dancing. The demographics of the recruited participants are:
\begin{itemize}
    \item Age: range from 21 to 44 years old, average $30.2 \pm 5.9$ years,
    \item Pronouns: 11 she/her, 11 he/him, 1 they/them\footnote{Multiple selections and adding alternative pronouns were allowed.},
    \item Dancing experience: on a 5-point scale ranging from no prior experience (1) to experienced (5): 3 had no experience (1), 9 were novice (2), 4 had taken beginner lessons (3), 4 regularly danced (4), and 2 were experts or dance instructors (5).
    \item Prior experience with physically interactive robots (referred to as robotics experience), on a 5-point scale ranging from no exposure (1) to programmer (5): 3 had no notion of them (1), 7 had seen them before (2), 4 had studied them (3), 4 had used them (4), and 4 had programmed them (5).
    \item General comfort with close proximity to a full-size humanoid robot like the REEM-C prior to the experiment: on a 5-point Likert scale from strongly uncomfortable (1) to strongly comfortable (5): 6 were neutral (3), 7 comfortable (4) and 9 strongly comfortable (5).
\end{itemize}

\begin{figure}
    \centering
    \includegraphics[width=0.6\linewidth]{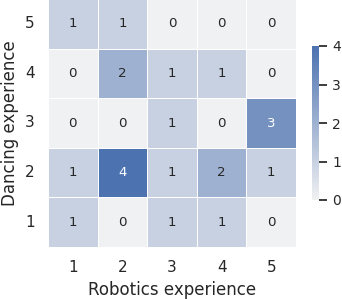}
    \caption{Prior participant experience distribution: each of the 22 participants is mapped with respect to their prior robotics and dancing experience. Experience is quantified on a 5-point scale as described in Sec.~\ref{sec:participants}. The number and shading of each box indicate how many participants belong in the corresponding intersection of experience. For example, 4 participants had seen physically interactive robots before (robotics experience value: 2) and were novice dancers (dancing experience value: 2).}
    \label{fig:pHRI_vs_dance_xp}
\end{figure}

Within the sample of participants gathered for this study, there appears to be a weak, negative correlation between prior dancing and robotics experience, Pearson's r = -.0958, N = 22; however, the relationship is not significant (p = .671), as illustrated in fig.~\ref{fig:pHRI_vs_dance_xp}. 

\subsection{Experiment design} \label{sub:experiment-design}
The different leading signals and their combinations are tested, given a randomized order of trials within three blocks, as shown in Table~\ref{tab:trials}. 
Each trial lasts $30~\si{s}$, during which the robot performs the sequence of steps and their accompanying leading signals in a loop. \edit{Commands to move the upper body, generate speech, and step, are sequentially sent to the robot in this order.}

Trials are run with the following values, \edit{determined through empirical tuning and confirmed with a pilot study}: 
\begin{itemize}
    \item When stepping, the distances traveled by robot feet are $0.13~\si{m}$ forward/backward, $0.145~\si{m}$ to the left/right,
    \item $ f^d = 1.5~\text{(virtual) } \si{N}$ for HW,
    \item $\delta^d= 5~\si{cm}$ for HD,
    \item $\Delta q_\text{torso} = 0.1~\si{rad}$ for TR, 
    \item The threshold on hand deflection $\boldsymbol{\widetilde{X}}_i$ is $0.15~\si{m}$, applied to the norm of the position vector (neglecting orientation),
    \item The threshold on wrench applied $\epsilon_\mathcal{F}$ is $\begin{bmatrix}
        5~\si{N} & 1.5~\si{Nm}
    \end{bmatrix}$, applied to the norm of the force and moment, respectively. 
    \item Minimum diagonal terms in $\boldsymbol{G}_J^i$  are set to 0.6 for torso joints, and to 0 for all other joints,
    \item $t_t = 500$ ms, i.e., transition time to gradually update gain values.
\end{itemize}

\begin{table}[tb!] %
    \centering
    \caption{Trials used in the participant experiments, separated into three blocks. The order of trials within each block is randomized.}

    \begin{tabular}{ p{3.7cm}   p{2.5cm}} 
        \textbf{Block} & \textbf{Trials}\\
        \hline
        
        \rowcolor{gray!20}
       \raggedright Block 1: \break 8 trials consisting of haptic/visual signals  & \makecell{
            NS \\ HW \\ HD \\ TR \\ HW + HD \\ 
            HW + HD2\footnotemark[1] \\
            HW + TR \\
            HW + HD + TR
            } \\
       \raggedright Block 2: \break 2 trials with audio signals  & \makecell{SC \\ SD}  \\
       \rowcolor{gray!20}
       \raggedright Block 3: \break 3 trials, combined haptic/visual and audio & \makecell{
            HW + SC \\
            HD + SC \\
            TR + SC
       }
    \end{tabular}
    \footnotetext[1]{Due to operator mistake, HW + HD was repeated, instead of HD + TR. While this results in a slight loss of information, the repetition allowed to assess the consistency of participant feedback on a given leading signal.}
    \label{tab:trials}
\end{table}

\subsection{Experimental protocol}\label{sec:experimental-protocol}

The experiments took place at the Human-Centred Robotics and Machine Intelligence Lab at the University of Waterloo (Canada). The experimenters welcomed participants into the room, and had the robot introduce the experiment given a prepared script that mentions safety, interactive controller features, and that the robot would be leading a dance. The participants were then asked to fill out a consent form, as well as demographic and pre-experiment questionnaires.

Afterwards, the experimenters demonstrated safe interaction with the robot, how to hold the robot's hands, and expected robot movements. The experimenters then adjusted the robot hand configurations for the participant's comfort, if required. This allowed to account for height differences, or how some participants preferred to connect with their palms facing up. 
Once the participant was ready to start with the experiment, one of the experimenters stated the number of trials, without specifying whether the robot would act differently between trials.

After each trial, the participant was asked to complete a trial questionnaire on a computer. Once all trials were done, the participant filled in a post-experiment questionnaire. In total, the experiment took 0.5 to 1 hour, depending on the level of detail in the participant's feedback. 

\begin{table}[htb!] %
    \noindent
    \caption{Questions asked of participants}

    \begin{tabular}{p{7.1cm}} 
        \hline
        
        \rowcolor{gray!20}
        \textbf{Demographics} \\
            What is your date of birth? \\
            What are your pronouns? \\
            What is your previous dancing experience? \\
            What is your previous experience with collaborative or physically interactive robots?\\
       \rowcolor{gray!20}
       \textbf{Pre-experiment} \\ 
            How comfortable are you about close proximity to a full-size humanoid robot like the REEM-C? \\
            How will dancing with the robot be different from dancing with someone else? \\
            How do you expect the robot to lead the dance? \\
            Do you have other thoughts? \\
       \rowcolor{gray!20}
       \textbf{After each trial} \\
            How well did you know what step to take? \\
            How comfortable did you feel during the dance? \\
            What did you like in particular? \\
            What did you dislike in particular? \\
            In this block so far, which trial felt the best? \\
            In this block so far, which trial felt the worst? \\    
       \rowcolor{gray!20}
       \textbf{Post-experiment} \\
            Between the three blocks of trials, which felt the best? \\
            Between the three blocks of trials, which felt the worst? \\
            What else would you have liked when dancing? Try to be specific. \\
            How could the interaction have been made better? \\
            How comfortable are you now about being in close proximity with a robot? \\
            Would you dance again with a robot in the future? \\
    \end{tabular}

    \label{tab:questionnaires}
\end{table}

The questions asked in each questionnaire are summarized in Table~\ref{tab:questionnaires}. 
Given the exploratory nature of the proposed experiment, and the lack of a suitable validated questionnaire (as discussed in Sec.~\ref{sec:related work}), questions were carefully formulated by the research team. Pre-experiment questions were selected to capture prior expectations from participants. The set of questions asked after each trial was defined to measure the impact of leading signals on participants' comfort and accomplishment of the dance steps. Open-ended questions were included to provide understanding of what aspects of the interaction were meaningful to the participants, without relying on prior assumptions.

Participants were asked for their preferred trial within the current block (instead of overall), after each trial. This choice was made in consideration of known working memory capacity limitations: (i) short-term memory is known to only retain several chunks of information~\cite{Cowan2001_memory_chunks}, and (ii) memory accuracy is known to decline with increasing memory load, with a sharper decline when information chunks share features~\cite{Oberauer2006_working_memory_capacity}. As a result, participants may not accurately retain information beyond a handful of trials, especially given that trials shared similarities (e.g., in participant and robot movements). This woud likely have led the participants' answers to be unreliable, if they had been asked ``Overall, which trial felt the best?''. For this reason, post-experiment, we asked participants to identify their preferred block, instead of trial. \edit{We acknowledge that this may lead to information not being fully captured.}

The remaining post-experiment questions were aimed at understanding what the experiment design may have missed but would be meaningful to participants in this psHRI scenario, as well as evaluating whether the experiment affected participants' expectations of psHRI.

With the experiment and questionnaires being developed as an early exploration of psHRI and human responses to it, validating the questionnaires was not considered at this point. 
\edit{As a result, measurement biases may be present in the reported results. For example, `comfort' is not explicitly defined, and may be interpreted in different ways by participants. What makes a trial better or worse is also subjective to each participant. The follow-up open-ended questions serve to capture participants' perspectives, but the possibility remains that they may not capture their complete perspective.}
A \edit{cautious and} critical analysis of participant feedback will help understand whether questions effectively captured participants' perception of the interaction, and will help extract insights for the design of future questionnaires adapted to psHRI.


\section{Experimental results} \label{sec:results}

\begin{figure}[bt]
    \centering
    \includegraphics[width=0.95\linewidth]{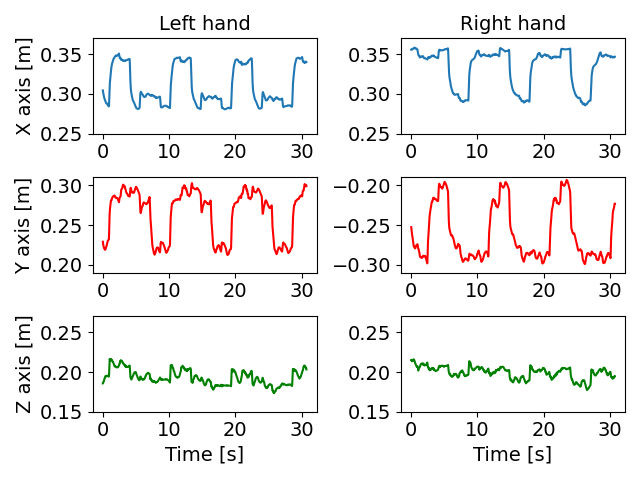}
    \caption{Typical left and right robot hand motion resulting from using the hand displacement (HD) leading signal during participant trials, expressed with respect to the base frame. The X, Y and Z axes point to the front, left and top of the robot, respectively. Differences in displacement between the left and right hands may be attributed to contact forces exchanged between the participant and the robot and variations in force-torque sensor calibration.}
    \label{fig:hand displacement}
\end{figure}

\begin{figure}[bt]
    \centering
    \includegraphics[width=0.95\linewidth]{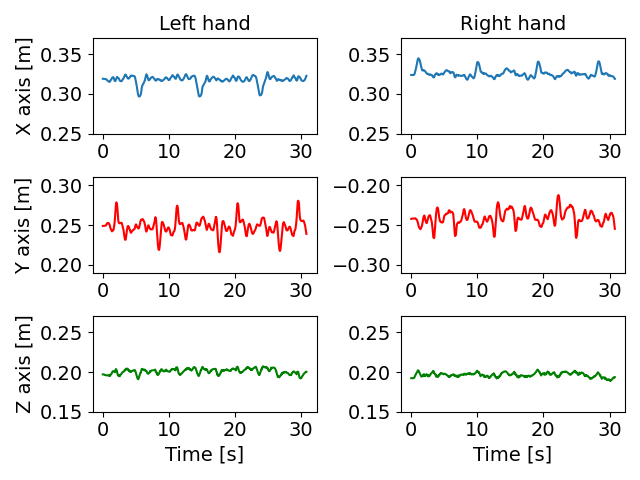}
    \caption{Typical left and right hand forces resulting using the hand wrench (HW) leading signal during participant trials. Note that the hand wrench is by definition virtual: the controller transforms the desired wrench in a corresponding compliant hand motion as shown here. The X, Y and Z axes are as defined in Fig.~\ref{fig:hand displacement} }
    \label{fig:hand wrench}
\end{figure}

\begin{figure}[bt]
    \centering
    \includegraphics[width=0.9\linewidth]{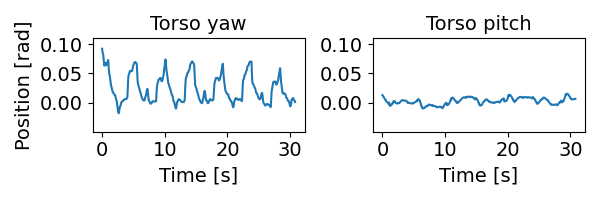}
    \caption{Typical torso rotation resulting from the TR leading signal during participant trials: the torso yaw is set to rotate with the steps, while the pitch is to remain constant.}
    \label{fig:torso rotation}
\end{figure}

Results that relate to the stated contributions of the paper are reported here. In all experiments, the robot functioned as expected: it led the interaction with the specified leading signals (for example as shown in Figs.~\ref{fig:hand displacement}, ~\ref{fig:hand wrench} and ~\ref{fig:torso rotation}), while remaining compliant to forces applied by a participant, and it would stop dancing if the participant pushed its hands away by more than the given threshold. The default balancing controller for the REEM-C can however make the robot vibrate at times; participants mentioned feeling these vibrations in their feedback. 
Nonetheless, this did not seem to negatively affect participants' general comfort towards interacting with the REEM-C\footnote{This \textit{general} comfort is not to be confused with the \textit{direct} comfort feelings experienced by participants during each trial. To disambiguate these two, the general comfort will be at times referred to as the general \textit{attitude} of participants.}.
Both the pre- and post-experiment questionnaires asked participants how comfortable they were about close proximity to a full-size humanoid robot like the REEM-C, on a scale from 1 (strongly uncomfortable)
to 5 (strongly comfortable).
Pre-experiment, answers averaged $4.1 \pm  0.8$ and post-experiment $4.5 \pm  0.6$; results of a t-test indicate a statistically significant increase pre- to post-experiment ($t = -4.73$, $p < .01$). Responses increased by 2 points for 1 participant, 1 point for 6, did not change for 14 participants, and decreased by 1 point for 1 participant (who moved from strongly comfortable to comfortable). None responded ``No'' when asked ``Would you dance again with a robot in the future?''.

\begin{figure}[tb]
    \centering
    \includegraphics[width=\linewidth]{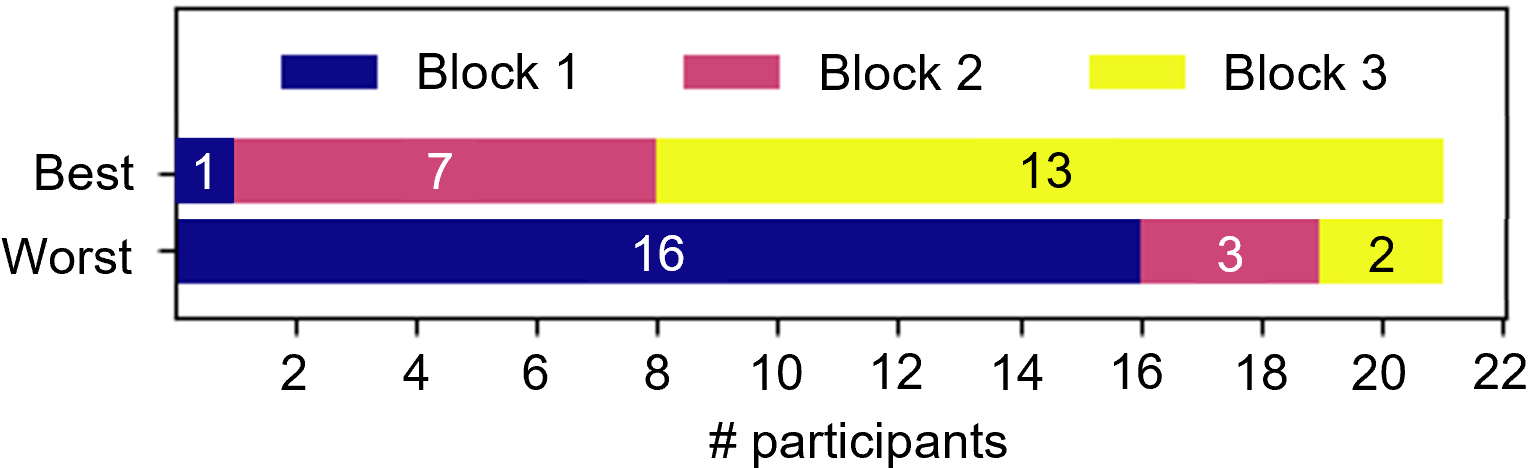} \vspace{-0.4cm}
    \caption{Distribution of blocks selected as best and worst. Note that one expert dancer abstained to answer, given that the robot behaved so differently from a human dancer.} 
    \label{fig:best-worst-blocks-distribution}  
\end{figure}

After the last trial, participants were asked ``Between the three blocks of trials, which felt the best and the worst?''. Results are shown in Fig.~\ref{fig:best-worst-blocks-distribution}:
13/21 \textbf{participants preferred block 3 (trials that combined audio with haptic/visual signals)}, 16/21 reported block 1 (only haptic/visuals signals) as worst, and one participant abstained to answer. A chi-square test of goodness-of-fit performed on participants' selections of best ($\chi^2(2, 21) = 10.3$, $p < .01$) and worst ($\chi^2(2, 21) = 17.4$, $p < .01$) blocks reveals a statistically significant difference.

\begin{figure}[tb]
    \centering
    \includegraphics[width=\linewidth]{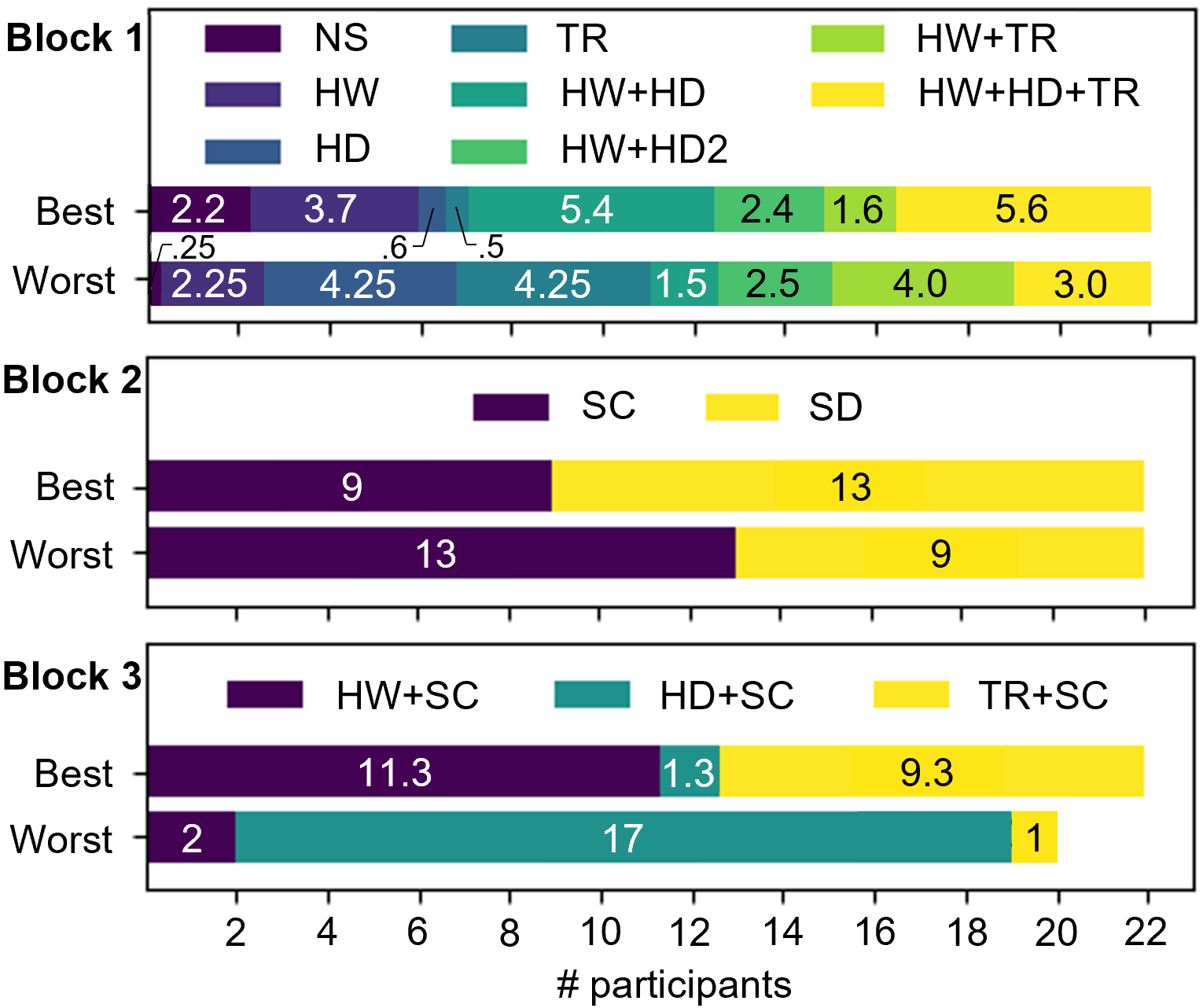}

    \caption{Distribution of trials selected as best and worst by participants, from top to bottom: block 1, block 2 and block 3. Multiple selections were allowed: e.g., if a participant selected $n$ trials as best, each of these trials is given a weight $1/n$. For block 3, two participants abstained from selecting a worst trial. 
    } 
    \label{fig:best-worst-trials-in-each-block}
\end{figure}

Within each block, after each trial, participants were also asked ``In this block so far, which trials felt the best and the worst?''. As shown in Fig.~\ref{fig:best-worst-trials-in-each-block}, for the first block, a clear consensus did not arise from participant feedback. TR and HD were selected the most often as worst, followed by HW + TR. For the best signals, HW + HD + TR and HW + HD were selected the most often, followed by HW. However, a chi-square test of goodness-of-fit reveals that the difference between best ($\chi^2(7, 38) = 7.05$, $p = .423$) and worst ($\chi^2(7, 28) = 5.14$, $p = .643$) selections is not statistically significant.

In the second block, 59\% of \textbf{participants preferred the steps to be described by the robot} over having the robot count the steps. This percentage was made up of 2/2 dance experts, 2/4 regular dancers, 1/4 beginners, 8/9 novices and 0/3 non dancers. Participant comments indicate that novice dancers appreciated instructions on where to step, whereas the expert dancers preferred the timing of the description audio with respect to when the stepping foot touched down. While a chi-square test of goodness-of-fit confirmed that the preference for the steps description over the step number was not statistically significant ($\chi^2(1, 22) = 0.727$, $p = .394$), a one-way ANOVA test confirmed that participants' dance experience significantly affected their reported preference ($F(4, 18) = 4.32$, $p = .0137$). Performing a Tukey HSD post-hoc test confirmed the presence of a statistically significant difference between participants who had no prior dancing experience and those who were novice dancers. Repeating the experiments with a larger sample of participants would help better assess the effect of higher levels of dance experience.

Block 3 revealed the clearest consensus: HD + SC, \textbf{using the hand displacement signal while counting steps, was least appreciated}. A chi-square test of goodness-of-fit confirmed the presence of a statistically significant difference between the selection of the three signals in block 3 as best ($\chi^2(2, 23) = 12.1$, $p < .01$) and worst ($\chi^2(2, 20) = 24.1$, $p < .01$).

\begin{figure}[bt!]
    \centering
    \includegraphics[width=\linewidth]{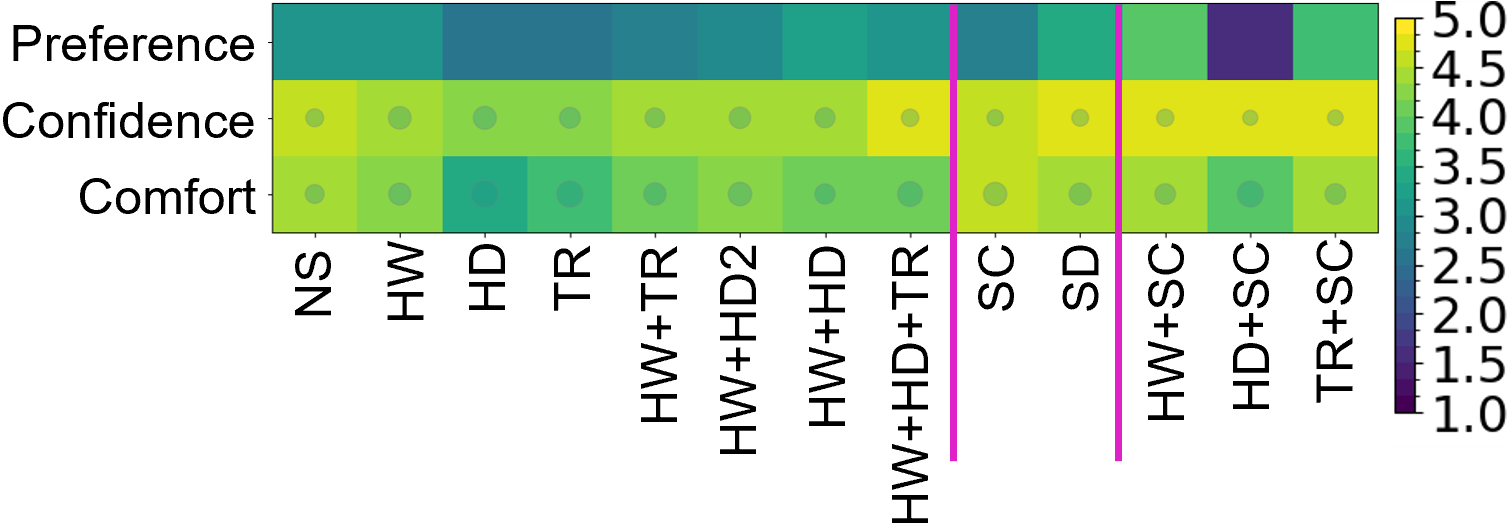} \vspace{-0.2cm}
    \caption{Average participants' level of comfort, confidence in doing the task, and \textit{preference metric} (eq.~\ref{eq:preference}) for each type of leading signal. The scale of the overlaid circles on the bottom two rows indicate standard deviation, ranging from 0.4 for the smallest circle, to 1.3 for the largest circle.}
    \label{fig:comfort_confidence_preference} 
\end{figure}

After each trial, participants were asked ``How well did you know what step to take?'' to get a measure of their confidence in understanding cues from the robot, and ``How comfortable did you feel during the dance?'' to see if changes in the robot's behaviour affected participants' comfort level. Both of these questions were answered on a Likert scale of 1 (No clue / Clearly uncomfortable) to 5 (Confident / Clearly comfortable). Responses averaged over all participants for each trial are shown in Fig.~\ref{fig:comfort_confidence_preference}, alongside a \textit{preference metric}\footnote{Know that while the preference metric $P_\text{trial}$ is helpful for a quick visual comparison/correlation assessment in Fig.~\ref{fig:comfort_confidence_preference}, it may not capture the full complexity of the situation.} $P_\text{trial}$ defined on a scale of 1 (all participants selected the trial as worst) to 5 (all participants selected the trial as best) calculated with:
\begin{equation} ~\label{eq:preference}
    P_\text{trial} = 3+ \frac{2}{n_p} \sum_{i = 0}^{n_p} \begin{cases}
        ~1 \text{\small{~~if }} i \text{\small{ selected the trial as best}}\\
        -1 \text{\small{ if }} i \text{\small{ selected the trial as worst}}\\
        ~0 \text{\small{~~otherwise}}
    \end{cases}
\end{equation}

One-way repeated measures ANOVA was conducted to evaluate the effect of leading signals on reported comfort and confidence. Results demonstrate that across all blocks, leading signals had a moderate statistically significant effect on comfort ($F(12, 252) = 3.86$, $p < .001$, $\eta_p^2=.085$) and confidence ($F(12, 252) = 2.26$, $p = .010$, $\eta_p^2=.071$). Post-hoc pairwise t-tests performed with Benjamini/Hochberg FDR correction identified statistically significant differences in comfort between HD and HW+HD2, HW+HD+TR, SC, SD, TR+SC, as well as between HD+SC and TR+SC. Within block 3, the difference in comfort between HW+SC and HD+SC, as well as between HD+SC and TR+SC was found to be significant.
Similarly, statistically significant differences in confidence were observed between middle trials in block 1 (HW, HD, TR, HW+TR) and HW+HD+TR plus trials in later blocks (SD, HW+SC, HD+SC, TR+SC).
Participants' worst and best trial selections are weakly correlated with their reported comfort (Pearson's $r= .268$, $p < .01$, $N=155$), and moderately correlated with their reported confidence (Pearson's $r= .318$, $p < .01$, $N=155$).

Among the first block trials, trial \textbf{HW + HD + TR yielded high confidence at the cost of comfort}. HD, used alone or with SC, yielded the lowest comfort, resulting in \textbf{HD + SC being the least preferred trial} overall. SC yielded the highest average comfort level, but with high standard deviation, and low confidence.
Thus, across all blocks, \textbf{HW + SC was the preferred trial}, with higher reported confidence and comfort. 

\begin{figure}[bt!]
    \centering
    \includegraphics[width=\linewidth]{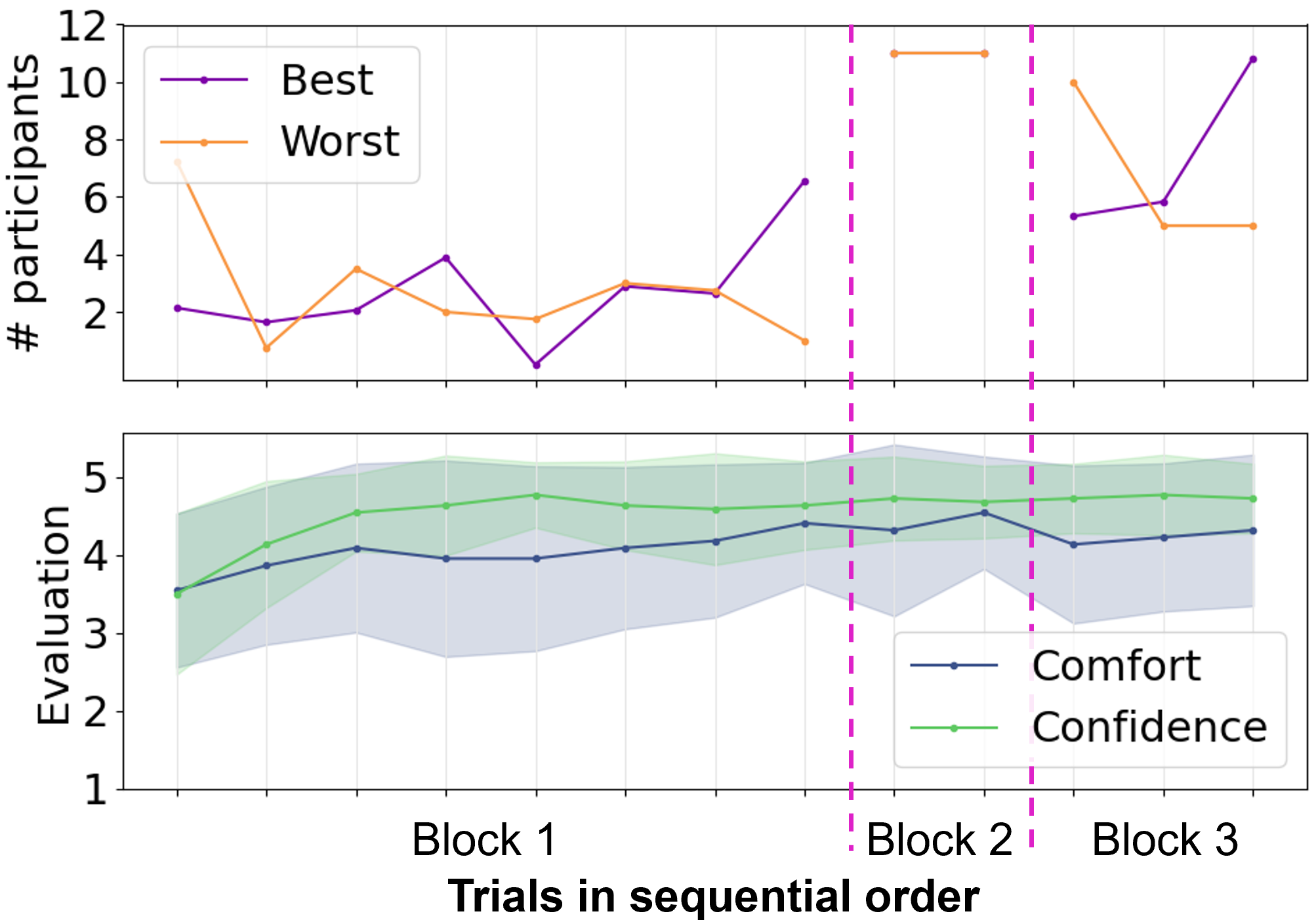}
    \caption{Top: number of trials selected as best and worst by participants, in function of their order (e.g., 11 participants selected the third trial of block 3 as best in that block. Bottom: evolution of participants' evaluation of their comfort and confidence, by trial number. Lines follow the average values and filled areas show the standard deviation.}
    \label{fig:best-worst-blocks-by-trial-number}
\end{figure}

To assess whether trial order mattered in participants' evaluation of trials, we plotted Fig.~\ref{fig:best-worst-blocks-by-trial-number}. It shows a possible bias towards the last trial as best, and the first as worst. Overall, there appears to be a significant strong positive correlation between trial order and the number of times a trial was selected as best (Pearson's $r = .751$, $p < .01$, $N = 13$), but only a non-significant moderate positive correlation for the number of times a trial was selected as worst (Pearson's $r = .443$, $p = .129$ $N = 13$). However, within block 1, there was no significant correlation between trial number and best ($r = .540$, $p = .167$, $N = 8$) or worst ($r = -.504$, $p = .203$, $N = 8$) trial selection.
On the other hand, within block 3, very strong but non-significant opposite correlations were found between trial number and frequency of selection of a trial as best ($r=.904$, $p=.281$, $N=3$) and worst ($r=-.866$, $p=.333$, $N=3$). 

\begin{figure}[tb!]
    \centering
    \includegraphics[width=0.9\linewidth]{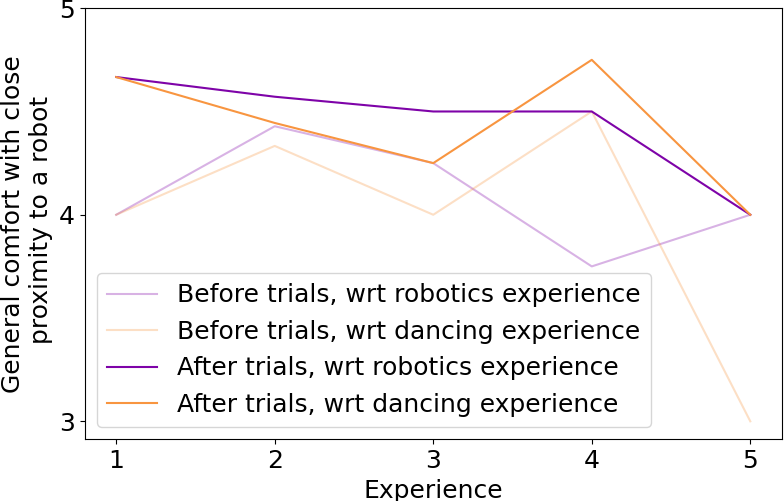}
    \caption{Influence of prior experience with physically interactive robots or with dance on the general attitude of participants towards close proximity to a robot, as reported before and after the experiment. `wrt' is short for `with respect to'.
    }
    \label{fig:background_vs_predicted_comfort}
\end{figure}

\begin{figure*}[htb!]
    \centering
    \includegraphics[width=0.49\linewidth]{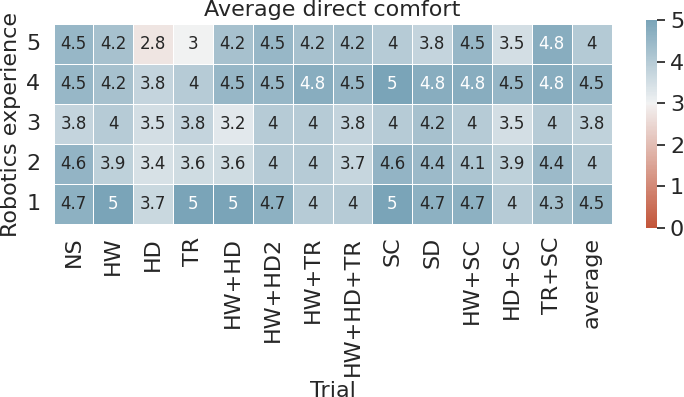}
    \includegraphics[width=0.49\linewidth]{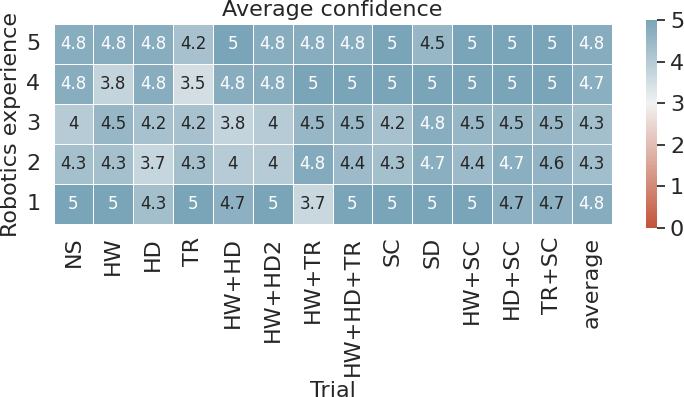}
    \caption{Influence of experience with physically interactive robots ($y$-axis) on the average direct comfort and confidence (heat  map values) reported for each trial ($x$-axis). For example, the participants who reported a level 4 experience reported an average confidence level of 3.5 during the TR (torso rotation) trial.}
    \label{fig:robotics_background_vs_comfort_and_confidence}
\end{figure*}

\begin{figure*}[htb]
    \centering
    \includegraphics[width=0.49\linewidth]{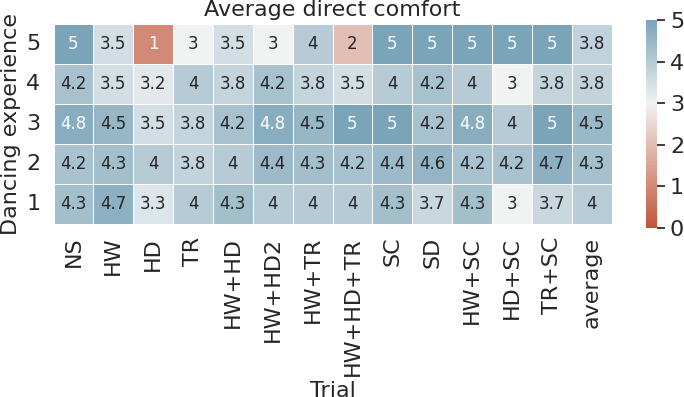}
    \includegraphics[width=0.49\linewidth]{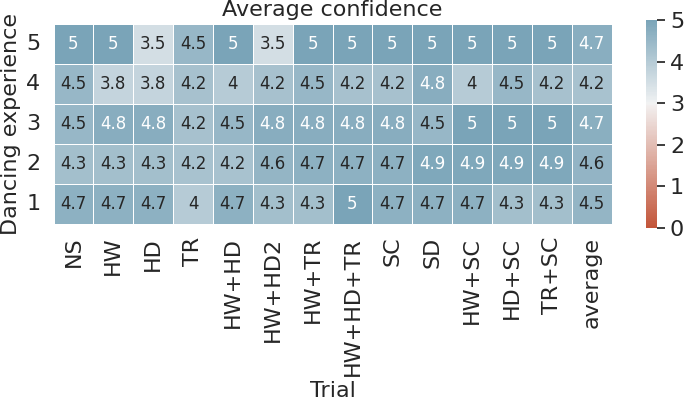}
    \caption{Influence of dancing experience ($y$-axis) on the average direct comfort and confidence (heat  map values) reported for each trial ($x$-axis). For example, the participants who reported a level 5 dancing experience reported an average direct comfort level of 1 during the HD (hand displacement) trial.}
    \label{fig:dancing_background_vs_comfort_and_confidence}
\end{figure*}

\begin{figure*}[htb]
    \centering
    \includegraphics[width=0.49\linewidth]{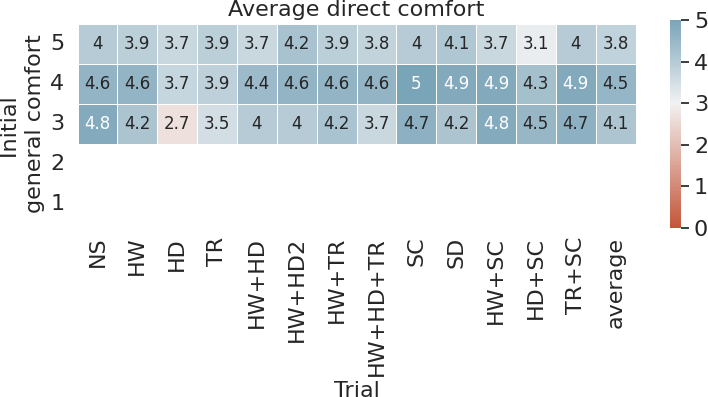}
    \includegraphics[width=0.49\linewidth]{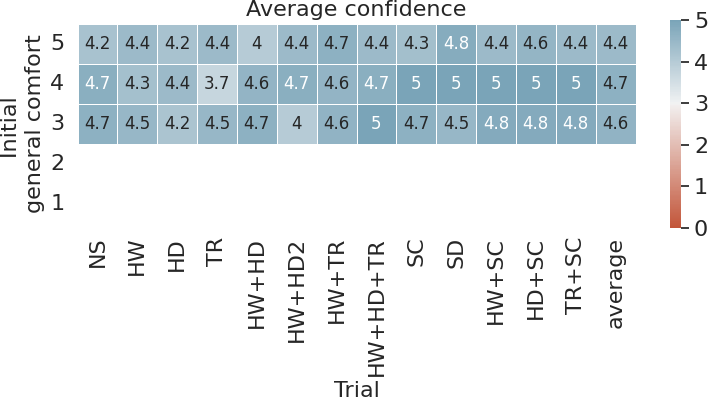}
    \caption{Influence of initial general comfort with close proximity to a robot ($y$-axis) on the level of direct comfort and confidence reported by participants (heat  map values) for each trial ($x$-axis). 
    For example, the participants who reported an initial general comfort of 3, in average reported a direct comfort level of 4.8 during the NS trial.}
    \label{fig:initial_vs_reported_comfort}
\end{figure*}

\begin{figure}[htb]
    \centering
    \includegraphics[width=0.49\linewidth]{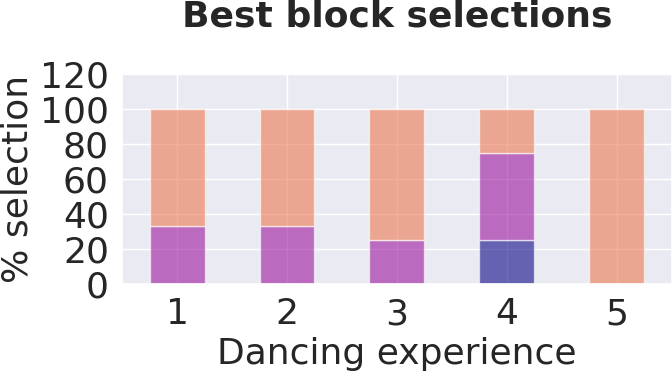}
    \includegraphics[width=0.49\linewidth]{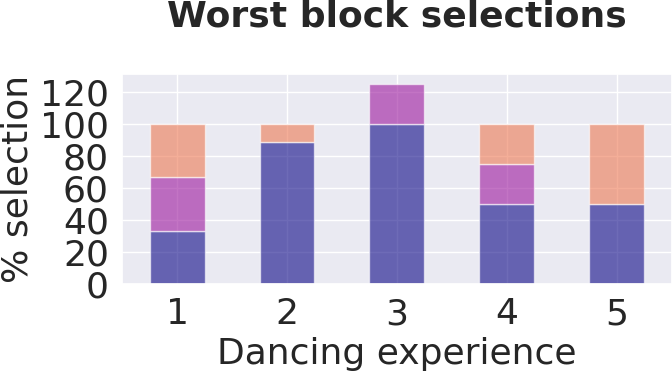}
    
    \includegraphics[width=0.49\linewidth]{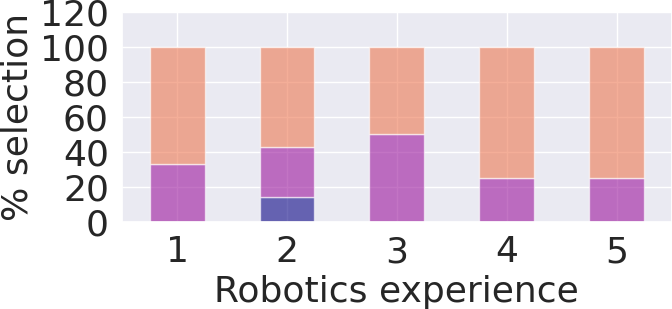}
    \includegraphics[width=0.49\linewidth]{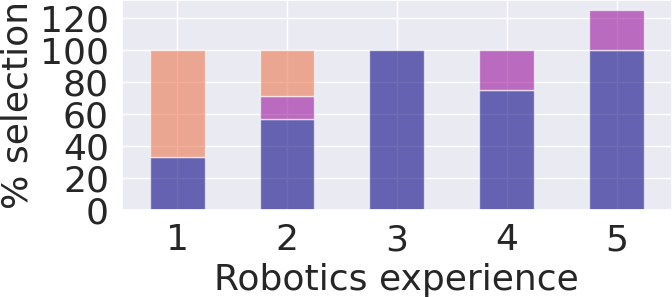}

    \includegraphics[width=0.49\linewidth]{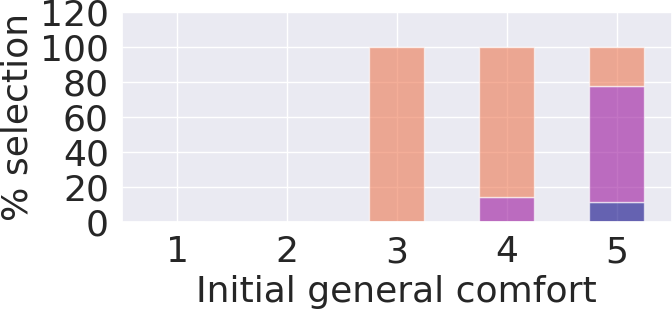}
    \includegraphics[width=0.49\linewidth]{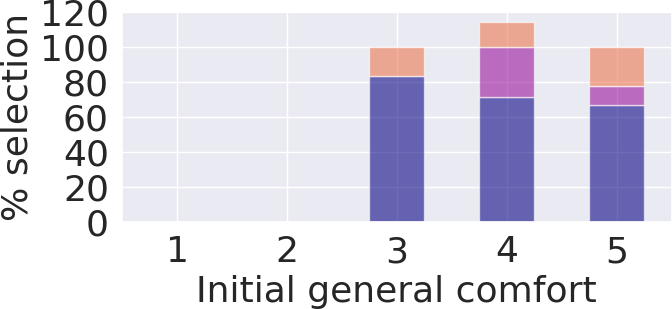}
    
    \caption{Influence of participant background on selection of best and worst trial blocks. Each graph shows the percentage of blocks selected as best (left) or worst (right), in terms of participants' dancing experience (top), robotics experience (middle) and initial general comfort with close proximity to robots (bottom). A percentage higher than 100 indicates that a participant selected more than 1 block.}
    \label{fig:background_vs_best_or_worst_block}
\end{figure}

One-way repeated measures ANOVA was conducted to evaluate the effect of trial order on reported comfort and confidence. Results demonstrate that trial order had a moderate statistically significant effect on comfort ($F(12, 252) = 2.44$, $p = .005$, $\eta_p^2=.057$), as well as a strong statistically significant effect on confidence ($F(12, 252) = 10.1$, $p < .01$, $\eta_p^2=.235$). 
Comparing with the effect of the leading signals themselves, trial order had a lesser effect on comfort and a larger effect on confidence. 
Post-hoc pairwise t-tests performed with Benjamini/Hochberg FDR correction identified statistically significant differences in comfort only between the first trial, and the 10th trial (i.e., the last trial of block 2). Similarly, statistically significant differences in confidence were only observed between the first trial and all other trials, as well as between the second trial and all other trials except the last three trials of block 1 and the last trial of block 2. 
Indeed, as Fig.~\ref{fig:best-worst-blocks-by-trial-number} shows, while \textbf{average confidence built up during the first few trials of the experiment and varied little afterward}, average comfort typically declined on the first trial of a block before slowly increasing in subsequent trials, and displayed a larger variance between participants. Overall, \textbf{the leading signals themselves had more effect on comfort than the order in which they were experienced}, and vice-versa for confidence.

We also investigated the possible effects of participant's background in dance or robotics on results. \edit{However, given the limited number of participants in this study, we may only provide exploratory observations, as opposed to statistically supported findings}. As shown in Fig.~\ref{fig:background_vs_predicted_comfort}, the two most experienced dancers reported the least level of general comfort with close proximity to a robot. A two-way ANOVA found no statistically significant effect of dance experience nor robotics experience on general comfort towards close proximity to robots, either pre- or post-experiment. 

Fig.~\ref{fig:robotics_background_vs_comfort_and_confidence} and \ref{fig:dancing_background_vs_comfort_and_confidence} explore whether participant background possibly affected the direct comfort and confidence participants reported during each trial. For instance, Fig.~\ref{fig:dancing_background_vs_comfort_and_confidence} shows that dance experts, who were generally confident they knew where to step, appeared to report the clearest discomfort during trials, but they also tended to agree on what trials felt the most comfortable. 
One-way ANOVAs found that participants' dancing experience had a statistically significant effect on direct comfort ($F(4, 281) = 4.42, p < .01 $), and confidence ($F(4, 281) = 4.02, p < .01 $). Post-hoc comparisons using the Tukey HSD test indicate that comfort was significantly different between (i) novice ($M=4.26$) and regular ($M=3.79$) dancers, as well as (ii) beginner ($M=4.46$) dancers and regular ($M=3.79$) or expert ($M=3.77$) dancers. Results also indicate significant differences in confidence between (i) novice ($M = 4.58$) or beginner ($M=4.71$) and regular ($M=4.23$) dancers, as well as (ii) regular ($M=4.23$) and expert ($M=4.73$) dancers. \edit{Nevertheless, these observations may be specific to the sample of participants for this study, and may not reliably generalize to a larger population.}

Similarly, participants' robotics experience had a significant effect on direct comfort ($F(4, 281) = 4.89, p < .01 $) and confidence ($F(4, 64) = 6.76, p < .01 $). Post-hoc comparisons using the Tukey HSD test indicate that comfort was significantly different between (i) participants who had no robotics experience ($M=4.51$) and those who had studied robotics ($M=3.83$), and (ii) participants who had seen a robot ($M=3.99$) or studied robotics ($M=3.83$) before, and those who had used a robot ($M=4.50$) before. They also indicate that confidence was significantly different between (i) participants who had no robotics experience ($M=4.77$) and those who had seen a robot ($M=4.35$) or studied robotics ($M=4.33$) before, as well as (ii) participants who had seen a robot ($M=4.33$) or studied robotics ($M=4.33$) before, and those who had used ($M=4.71$) or programmed ($M=4.79$) one.

\edit{These observations suggest that prior dance and robotics experience affected} how one experienced the interaction, likely based on prior expectations. 
For instance, 
\edit{in this study}, those who had studied robotics ended up experiencing the trials as more comfortable than would have been indicated from their initial general attitude towards close proximity to a robot, whereas beginner dancers instead experienced the trials as less comfortable than their initial general attitude would have indicated. 
Indeed, the participants whose initial general attitude indicated strong comfort around robots
turned out to report the least level of direct comfort during trials, as evidenced in Fig.~\ref{fig:initial_vs_reported_comfort}. A one-way ANOVA indicates that initial general attitude had a statistically significant effect on direct comfort ($F(2, 283) = 11.74, p < .01$), and confidence ($F(2, 283) = 3.656, p = .027$). Post-hoc comparisons using the Tukey HSD test indicate that comfort was significantly different between (i) participants who had a neutral attitude ($M=4.11$) and those with a positive attitude ($M=4.52$), as well as (ii) participants who had a positive attitude ($M=4.52$) and those who reported a strongly positive attitude ($M=3.84$). On the other hand, a statistically significant difference in confidence was observed between participants with a positive attitude ($M=4.67$) and those with a strongly positive attitude ($M=4.42$).
These results indicate that initial attitude also influenced participants' prior expectations of the interaction: a highly positive attitude might have translated into unrealistic or unmet expectations of the robot's abilities.

Fig.~\ref{fig:background_vs_best_or_worst_block} suggests that participant background may have influenced the selection of a preferred block: those who regularly or never danced appear more ambivalent as to which block they preferred or disliked most. Those with the most robotics experience disliked block 1 the most, while those who initially expressed a more positive general attitude towards close proximity to robots actually
most
disfavored blocks with haptic leading signals. One-way ANOVA \edit{indicates that, among the participants in this study,}
 general attitude had a statistically significant effect on participants' selections of best blocks ($F(2, 18) = 8.16, p = .003$) and worst blocks ($F(2, 18) = 3.90, p = .0392$), whereas dancing and robotics experience did not.
In particular, post-hoc comparisons using the Tukey HSD test indicate a significant difference between participants with a neutral or positive attitude, who favoured block 3, and those with a strongly positive attitude, who instead favoured block 2.

Note that while the intersection of prior dance and robotics experience may have influenced individual selections of best and worst trials, the distribution of participants' experience (as shown in Fig.~\ref{fig:pHRI_vs_dance_xp}) did not support the extraction of statistically relevant results, given the relatively small number of participants and large number of options. Therefore, our exploratory analysis will not go to such a granular level.

\begin{table}[bt!]
    \centering
    \caption{Participant feedback for each trial: number (\#) of participants who mentioned liking or disliking \handmotion{hand motion}, \torsomotion{torso motion}, \audio{audio}, or \other{other or unspecified aspects} of each trial, followed by words they used. Fig.~\ref{fig:best-worst-trials-in-each-block} may be referred to for the total number of participants who selected each trial as best or worst.
    }

    \begin{tabular}{ p{1.1cm} | p{2.6cm}   p{2.6cm}}
    \textbf{Leading} & \textbf{Liked}  & \textbf{Disliked} \\
    \textbf{signal(s)} & (\#)  Description & (\#) Description \\

    \hline
    
    \rowcolor{gray!20}
    \textbf{NS}  
    & \raggedright\handmotion{(1) smooth, better} \break \other{motion was smooth}
    & \handmotion{(2) not moving, no force} \\

    \textbf{HW} 
    & \handmotion{(6) gentle, nicer, subtle} 
    & \handmotion{(1) lack of engagement} \\

    \rowcolor{gray!20}
    \textbf{HD} 
    & \handmotion{(4) cool}
    & \handmotion{(11) aggressive, jerky, strong, sudden, uncomfortable, unexpected, weird} \break
    \other{not always giving clear signals}\\

    \textbf{TR} 
    & \handmotion{(1) no aggressive motion} 
    & \handmotion{(2) no motion} \\ 
    & \torsomotion{(1) unnatural}\\
    & \other{the sidestep was clearly feelable, steps were slightly clearer, the robot felt more careful} 
    & \other{felt slightly less assertive} \\

    \rowcolor{gray!20}
    \raggedright\textbf{HW + HD} 
    & \handmotion{(7) awesome, better, intuitive, nice} \break
    \other{very smooth and clear directions, the leading was gentle, less intimidating, more natural}
    & \handmotion{(3) heavy handed, jerky} \\

    \raggedright\textbf{HW + HD 2} 
    &\handmotion{(6) cool, dance-like, engaging, forceful, nice, less rigid}
    & \handmotion{(2) robotic, weird, less natural} \\
    & \torsomotion{(1) motion } \\
    & \other{smoother, felt more assertive in leading} \\ 

    \rowcolor{gray!20}
    \raggedright\textbf{HW + TR} 
    & \raggedright\handmotion{(3) helpful, responsive, slight, subtle} \break 
    \other{motion was slow and gentle }
    & \raggedright\handmotion{(1) jerky} \break \other{movements felt very disconnected from each other, no communication or engaging aspect}\\ 

    \raggedright\textbf{HW + HD + TR} 
    & \raggedright\handmotion{(8) adaptive, better, comfortable, less forceful, less jerky, natural} \break \torsomotion{(1) motion}
    & \raggedright\handmotion{(3) exaggerated, strong, too much} \break \break \torsomotion{(3) odd, intimidating} \\ 
    
    \end{tabular}
    \label{tab:lead-signal-feedback}
\end{table}
\begin{table}[tb!]
    \centering
    \caption{Continuation of Table~\ref{tab:lead-signal-feedback}} 
    \label{tab:lead-signal-feedback2}

    \begin{tabular}{ p{1.1cm} | p{2.6cm}   p{2.6cm}}

    \textbf{Leading} & \textbf{Liked}  & \textbf{Disliked} \\
    \textbf{signal(s)} & (\#)  Description & (\#) Description \\

    \hline
    
    \rowcolor{gray!20}
    \raggedright \textbf{SC} 
    & \raggedright\handmotion{(1) no motion} \break \audio{(14) make it easier to follow}
    & \raggedright\handmotion{(1) no motion} \break \audio{(9) the timing is off, numbers are not helpful} \\ 

    \textbf{SD} 
    & & \handmotion{(2) no motion}\\
    & \audio{(16) helpful when one does not know the steps} 
    & \audio{(9) condescending, missing what foot to move, too many words, not leading }\\

    \hline

    \rowcolor{gray!20}
    \raggedright\textbf{SC + HW}  
    & \raggedright\handmotion{(6) assertive, better, not forcing, gentle, slight} \break \audio{(2) timing felt better} \break \other{(4) combined arm motion and verbal instructions;} \break \other{(2) smoother, more natural}
    & \raggedright\handmotion{(1) unclear leadwork} \break \break \audio{(6) timing is off}\\ 

    \raggedright\textbf{SC + HD} 
    & \raggedright\handmotion{(5) very clear, more force} \break \audio{(1) audio} \break \other{(2) combined arm motion and verbal instructions}
    & \raggedright\handmotion{(14) aggressive, `being thrown around', jerky, heavy, rigid, sudden, too much, uncomfortable, violent} \break \audio{(2) unhelpful, wrong timing} \\

    \rowcolor{gray!20}
    \raggedright\textbf{SC + TR} 
    & \raggedright\handmotion{(5) no jerky motion} \break \torsomotion{(1) slightly moving} \break \audio{(5) timing felt better, more natural} \break \other{(2) feeling more comfortable}
    & \raggedright\handmotion{(5) no clear signal, less engaging, not leading} \break \audio{(4) timing is off}\\
    \end{tabular}
    \label{tab:lead-signal-feedback-continued}
\end{table}

After each trial, participants were asked for qualitative feedback with ``What did you like in particular?'' and ``What did you dislike in particular?''. We found that some participants tended to evaluate trials from an ego perception (e.g., ``I think I got better in following the robot's moves'', ``I am feeling more comfortable with the robot''), while others tended to evaluate how the robot behaved. Participants were also inclined to mix in general comments about the robot and the experiments (e.g., ``The robot’s motion is shaky''), among their feedback on a given trial. After isolating ego perception and general comments from trial-specific ones, we assembled the feedback on leading signals in Tables~\ref{tab:lead-signal-feedback}-\ref{tab:lead-signal-feedback2}.
For completeness, the remaining comments are reported in Appendix~\ref{sec:AppendixA}.

As can be glanced from Tables~\ref{tab:lead-signal-feedback}-\ref{tab:lead-signal-feedback2}, only trials including HW were qualified by participants with the word `gentle', while trials including HD were more typically qualified as `aggressive'.

Overall, \textbf{participants preferred having gentle haptic cues}: it helped them know where to step without looking at the robot's feet. Others mentioned that \textbf{being guided by voice positively affected the experience}. 


\section{Discussion and conclusions}\label{sec:conclusion}

The proposed control architecture
enabled the robot to lead partnered dancing through various haptic, visual and audio cues, resulting in varying perceived effectiveness and comfort in people, as demonstrated in participant feedback. 

\edit{While the comfort and confidence scores from the questionnaire need to be interpreted with caution due to the questionnaire not having been formally validated, the results demonstrate that the}
 questionnaire successfully elicited feedback from participants on what aspects of the robot's behaviour made them feel more of less comfortable, and on their progress learning the steps alongside the quality of the communication from the robot.
\edit{As shown in Sec.~{\ref{sec:results}}, leading signals and familiarity with the task both contributed to reported participant comfort and confidence, but familiarity may not have been the overall dominant factor driving results.
The results indicate that average participants' confidence with the steps may initially have been influenced by time and repetition more than leading signals, then plateaued around the middle of block 1. Block 1 results would thus be expected to reflect that participants were initially building familiarity with the task.
In terms of comfort, the analysis of the results over all three blocks showed that while average comfort slowly trended upward within each block of trials, the leading signals had a direct effect on participants' reported comfort during a trial. In addition, a number of participants reported making use of the leading signals to synchronize their steps with the robot's, as reported in Sec.~{\ref{sec:results}} and further detailed in Appendix~{\ref{app:lengthier_feedback}}.}

\edit{When interpreting results, one may also want to keep in mind that familiarity with the leading signals themselves may have affected reported comfort and confidence, such that the effects of familiarity with both steps and leading signals may not be fully separated.
While block 1 introduced new haptic and visual communication modalities, and block 2 introduced audio communication modalities, Block 3 instead combined modalities that participants had previously experienced. When comparing the effectiveness of leading signals between blocks, one may want to take into account that participant feedback would in part reflect whether they were adapting to, and learning to interpret, a new communication behaviour from the robot.}

\edit{As described in Sec.~{\ref{sub:experiment-design}}, not all possible combinations of individual leading signals were tested. For instance, to keep the duration of the experiment reasonable for participants, trials did not include the combination of SD and haptic/visual signals. The combination of HD and TR, however, was excluded by mistake. Even if such omissions impact the completeness of comparisons among communication behaviours, we believe that they may not have resulted in critical information being missed. While each trial is distinct from the others, we do not expect the participants' experience of a given signal combination to be so critically different as to change their overall experience if the combination was omitted. For example, knowing that (i) HD alone resulted in the worst average reported comfort and confidence, (ii) TR alone resulted in the second worst average reported comfort and confidence, and (iii) HD and TR were equally selected as the worst trials of block 1 by participants, if HD+TR had been included as a trial, it could potentially have been among the worst trials (if not the worst), but it would not be expected to be among the better ones. Similar deductive reasoning could be followed for SD+HW/HD/TR. 

Much more remains to be explored in terms of psHRI and human-robot communication behaviours.}
\edit{For one thing, the timing relationship between the steps and the haptic, visual and audio cues may directly affect participant perceptions of the interaction. In future work, the timing of the leading signals would need to be more carefully evaluated. For instance, combining haptic/visual cues with audio cues appeared to help some participants synchronize better with the robot's steps, but others still found the timing to be wrong. 
This timing issue is likely due to the programming of the robot being sequential: in block 1, the robot executed the command to generate a haptic/visual signal, and started moving its foot exactly as the upper-body motion ended. In block 2, the robot completed the speech command, before moving its foot. In block 3, the robot was programmed to start with the haptic/visual signal, proceed with the speech, and then step. As a result, the delay between the beginning of the leading signal and the step increased, from block 1 to block 3. While the compliance of the participant's body to the robot's motion could have dampened the participants' perception of the delay between the haptic cues and the steps, this was not the case with the visual and audio signals.
Further work on the temporal coordination of robot communication and actions during psHRI would help uncover deeper, more meaningful insights.}

Feedback from participants also demonstrates the need to consider, as part of psHRI deployment and evaluation, the quality of a robot's movements in terms of smoothness, rigidity, human/machine-likeness, responsiveness, assertiveness, predictability, and interpretability. 
Additionally, \edit{the exploratory observations from this study indicate that} participants' familiarity with robots and prior expectations are meaningful in the evaluation of psHRI. Together, these insights may be used in the future development of validated questionnaires adapted to psHRI.

Results obtained using the proposed questionnaires nonetheless demonstrate that combining compliance and assertiveness in robot movements, as with a combined admittance and impedance controller, \edit{can contribute} to comfort and clear communication in psHRI. 
Furthermore, trials that combined haptic/visual and audio signals \edit{appeared to make} it easier and more comfortable for participants to follow robot steps, whether consciously or unconsciously. Thus, we may recommend using multiple concurrent communication modalities in psHRI. Doing so would furthermore be in line with universal design principles, allowing communication between a robot and users with diverse (dis)abilities. 

It may also be appropriate to investigate how assertiveness is perceived in different psHRI scenarios, in order to find parameters that are conducive to comfortable, yet effective communication (e.g., in terms of motion speed and acceleration, forces exchanged, audio volume and tone...).
For instance, haptic signals should not be exaggerated or insufficient, lest individuals prefer doing away with them.  
Table~\ref{tab:guideline_table} provides a summary of the main communication guidelines which were extracted from the experimental results.

 \begin{table}[!tb]
        \centering
        \caption{Summary of guidelines for interpretable and comfortable communication in psHRI, extracted from experimental results}
        \begin{tabular}{p{7cm}} 
            
            \rowcolor{gray!20}           
            Gentle haptic cues help communication when in close proximity. \\
            
            Multi-modal communication is generally preferred (for example, combining haptic, audio, visual cues). \\

            \rowcolor{gray!20}
            Clear communication needs to be balanced with human comfort (for example, strongly assertive haptic cues may be more clear but less comfortable than gentle haptic cues). \\
            
            Non-experts in a task may prefer having the robot describe the task to them as they go through it; experts may prefer a robot to conform to their expectations of how to perform the task. \\

            \rowcolor{gray!20}
            Confidence and comfort in collaboratively carrying out a task with a robot are likely to increase with exposure, but comfort can momentarily drop when a new change is perceived. \\

            General attitude towards robots influences perceived comfort during psHRI (for example, an individual who tends to be unwary of close interactions with robots may perceive a given interaction as less comfortable than someone who is generally cautious about interacting with robots). \\

            \rowcolor{gray!20}
            Robot compliance combining impedance and admittance can support comfort and communication, but may need additional attention to establish a physical connection between human and robot.
        \end{tabular}
        \label{tab:guideline_table}
    \end{table}

Across experiments, torso rotation, hand displacement and force were set at constant values. 
Feedback revealed that some perceived the torso rotation as unnatural, the hand displacement too high, or the hand wrench a little low, resulting in TR unconsciously affecting comfort, HD feeling jerky, and HW feeling weak to some.  
Additional experiments allowing participants to tune controller parameters to their preference would provide deeper understanding of the stiffness/compliance, forces and displacements that yield a comfortable and intuitive interaction. 

This work assumed that one fixed set of control and signal parameters will work equally across individuals. 
\edit{While preliminary tests were conducted to tune parameters to reasonable values}, variation of parameters may affect people's perception of psHRI.
\edit{In addition, acknowledging the natural human diversity, different individuals will perceive psHRI differently. The present study, being of a limited scope, is not expected to have reliably captured and characterized these differences.}
As suggested in \cite{Abe2024PerceptionpHRI}, variance between participants is likely to be related to individual personality and background, \edit{of which the different expectations of an expert dancer versus a non-dancer, or of an expert in robotics versus not are only two of the many possibilities that could be explored}. Future work may seek to uncover more of these relations, and how robot behaviour may be adapted both during robot development and operation, to accommodate individual preferences and abilities.

Among other underlying assumptions of this work, we did not include consideration of language barriers, and for the most part, of sensorial and mobility impairments.
It can be expected that people with diverse abilities will experience the interaction differently.
Other typical aspects of dancing were also neglected, with limited engagement with the human partner: the robot's gaze did not follow them or scan the environment;
the robot did not provide any feedback to its follower or establish a proper connection in dancing terms.

On this note, an interesting insight stems from the use of a threshold on hand displacement to stop the robot dancing. While this event was not once triggered deliberately during participant experiments, we noticed a small number of participants had a tendency to push on the robot's hands while dancing. In particular, one participant with a jazz dance background expected the robot to mirror the pressure they exerted on the robot's hands. While the development of such behaviour was not included in this study, it may be beneficial to incorporate it in future dancing robot designs. In fact, motion controllers that can intelligently modulate compliance and resistance when completing a given task would also find use in other applications of physical human-robot interaction such as robotically-assisted rehabilitation, kinesthetic teaching, and co-manipulation.

Future work may also investigate the ease of interpretation and cognitive load associated with different robot communication behaviours, in various psHRI scenarios.

Nonetheless, while some may express scepticism about close physical interaction with robots, advances in psHRI, and particularly dance-based psHRI, hold promising implications for human assistance, physical therapy and psychotherapy~\cite{Feil-Seifer2005Defining, basso2021dance, Zhang2024DanceMovementTherapy}. 
The work presented above, in which participants held hands and danced with a full-size humanoid robot, demonstrates that individuals today are positively willing to engage in psHRI. 

Towards ensuring safety in a time where humanoid and non-humanoid robots are integrating workplaces, public spaces, and homes, often without appropriate standards and regulations in place~\cite[Chapter~7]{Diana2024-Moxi-design-book},~\cite{Prather2025-IEEEHumanoidReport, Prather2026-IEEERobotsInPublicReport, Laursen2026-ISO-home-robots}, it is critical to understand how humans will spontaneously approach robots, as well as how to design and establish communication between humans and robots.
As such, the work presented above consists in a fundamental step in 
improving the ability of robots to
communicate with humans through haptics, posture and audio,
towards making physical human-robot collaboration safe, comfortable and intuitive. On the long term, our work will support progress towards robots that can be deployed in collaborative applications and effectively work with people who have varying levels of robotics experience, inclusive of those who have minimal training with robots.


\backmatter

\bmhead{Acknowledgements}

The authors would like to extend special thanks to Vidyasagar Rajendran for his feedback on the controller, and to all participants for their valuable contributions.

\section*{Declarations}

\subsection{Ethics approval}

This study has 
received ethics clearance through the University of Waterloo
Research Ethics Board (REB 43485). 

\subsection{Data availability statement}
The video, robot and questionnaire data that support the findings of this study are available upon reasonable request from the corresponding author.

\subsection{Consent to participate}
Informed consent was obtained from all individual participants included in the study. 

\subsection{Consent to publish}
The authors affirm that participants provided informed consent for publication of the images in Figure 1 and the video attachment. 

\subsection{Funding}
This work was supported by the Natural Sciences and Engineering Research Council of Canada, the Tri-Agency Canada Excellence Research Chair Program, the University of Waterloo and the Hector Foundation (Germany).

\subsection{Competing interests}
The authors declare they have no relevant financial or non-financial interests.




\bibliography{bibliography}

\begin{appendices}

    \section{Additional participant feedback}\label{sec:AppendixA}

This appendix reports on other feedback provided by participants which was not reported in the main body of the paper. This includes prior expectations of participants, additional feedback provided during the trials, as well as post-experiment feedback including recommendations on how to improve the interaction.

\subsection{Prior expectations of robot dancing} \label{sec:Appendix_prior_expectations}

Before beginning the trials, participants were asked about what they expect from dancing with the robot through three questions:
\begin{enumerate}
    \item ``How will dancing with the robot be different from dancing with someone else?''
    \item ``How do you expect the robot to lead the dance?''
    \item ``Do you have other thoughts?''
\end{enumerate}
Freeform responses from participants are summarized in Tables~\ref{tab:prior_expectations_dancing}, \ref{tab:prior_expectations_leading} and \ref{tab:prior_expectations_other_thoughts}.

 \begin{table}[!b]
        \centering
        \caption{Summary of answers provided by participants when asked ``How will dancing with the robot be different from dancing with someone else?''. (\#) reports the number of participants who mentioned a specific aspect when that number is more than one.}
        \begin{tabular}{p{2cm}   p{4cm}}
            \textbf{Topic} & \textbf{Detailed description} \\
            \hline
            \rowcolor{gray!20}
            Motion & Rigid (6), less reactive (6), slow (4), less flow (2), preprogrammed, jerky \\
            
            Interaction & Unusual connection (2), no emotional interaction (2), no eye contact, no body language, no facial expression, different feeling of dancing \\
            \rowcolor{gray!20}
            Safety & Different\\

        \end{tabular}
        \label{tab:prior_expectations_dancing}
    \end{table}

     \begin{table}[!b]
        \centering
        \caption{Summary of answers provided by participants when asked ``How do you expect the robot to lead the dance?''. (\#) reports the number of participants who mentioned a given aspect.}
        \begin{tabular}{p{6cm}}
            \rowcolor{gray!20}
            Through hands, using touch (5) or motion (2) \\
            Rigidly following a trajectory (4) \\
            \rowcolor{gray!20}
            Slowly (2) \\
            Through voice (1) \\
            \rowcolor{gray!20}
            The same as in partner dancing (1) \\
            No idea (8) \\
        \end{tabular}
        \label{tab:prior_expectations_leading}
    \end{table}

\begin{table}[!tb]
        \centering
        \caption{Summary of answers provided by participants when asked ``Do you have other thoughts?'' prior to the experiment. (\#) reports the number of participants who mentioned a thought, if more than one.}
        \begin{tabular}{p{6cm}}
            \rowcolor{gray!20}
            Will the robot step on my toes? (2) \\
            The robot may be slow (2)\\
            \rowcolor{gray!20}
            This is super cool. I am excited!\\
            This will be interesting. I'm as interested in what does not work as I am in what does work.\\
        \end{tabular}
        \label{tab:prior_expectations_other_thoughts}
    \end{table}

\subsection{General and self-perception comments provided during the trials}

After each trial, participants were asked ``What did you like in particular?'' and ``What did you dislike in particular?''. Answers that were relevant to each trial have been presented in the main text. The remainder have been sorted between general and self-perception comments.

Below are general comments which were not necessarily tied to any specific trial, with the total number of times a certain aspect was evoked between brackets:
\begin{itemize}
    \item The robot's motion is shaky (33), rigid (6), mechanical (4)
    \item The movements are consistent, repetitive and predictable (18)
    \item The steps are small (6)
    \item The pace is nice (3)
    \item The robot's weight shifts are fairly clear (2), unclear (1)
    \item The upper-body is responsive to one's movements while still showing the steps (2)
    \item There is a sudden jitter of the robot at the start of each trial (1)
    \item The robot is patient (3), careful (2), quiet (2), slow (1)
    \item The robot's voice is not like that of a human (1), is annoying (1), sounds angry or bossy (1), sounds academic (1), sounds robotic (1)
    \item The robot's motors make noise (3); they can serve as audible cues (1)
    \item The robot's gaze is not directed at the partner's face (1)
    \item The eyes of the robot are cold black dots (1)
    \item The voice feedback made me have more contact with the robot's face (1)
    \item ``Overall, most of the things are alright'' (1)
\end{itemize}

Self-perception comments mostly consisted of reflections on a participant's progress learning the dance, and can be summarized as follows:
\begin{itemize}
    \item I am following the robot more easily (10)

    \item I am feeling more comfortable with the robot (9)

    \item I feel confident with respect to the robot's movement and balance (4)

    \item I am feeling more familiar/confident with the dance (4)

    \item I looked down at the robot's feet less (3)
        
    \item I have been looking at the legs to know what to do (1)
    \end{itemize}

Other self-reflective comments were about specific aspects that a participant liked:
\begin{itemize}
    \item ``I tried pushing the robot back a little so it was looking more at my face. That was nice.''
    \item ``I liked the feet movement''
    \item ``I like the fact that it is a robot''
    \item ``I like that I am dancing with a robot''
    
\end{itemize}

And the remaining are comments on aspects that a participant disliked, with common themes on the physical and visual interaction, as well as perceived stress during the experiments:
\begin{itemize}
    \item Physical and visual interaction
    \begin{itemize}
         \item ``I was not sure exactly how to put my hands in a comfortable position because the hands of the robot are rigid and bigger than my own.''
        \item ``I think holding the hands of the robot is still something odd. I would not say that I felt a personal connection that I think is needed in this type of dancing. This time I tried to look at its neck/chest instead of the face because I also still think looking into void eyes is a bit odd.''
        \item ``I need to figure out my hand positioning with respect to the compliance receptors and how to avoid unintentionally applying too much force.''
        \item ``If this robot asked me to dance, knowing what I know, I would say no because the dance is not enjoyable.''
    \end{itemize}
    \item Perceived stress
    \begin{itemize}
        \item ``The robot seemed a little overconfident in its big steps to the side, so I felt like I had to compensate for those steps by holding it back more than in the rest of the dance.''
        \item ``In the previous trial, the arm tugging was rather intense, and to prevent my arms from being tugged away intensely, I was more reluctant to move my arms with the robot.''
        \item ``I find it stressful to try to follow, since follows are meant to match the tone, and I am not a robot.''
        \item ``I stress that something unexpected might happen''
        \item ``The robot doesn't appear to be learning anything from one trial to the next or speeding up at all, so it feels like I'm doing more work or putting in more effort than the robot is''
     \end{itemize}
\end{itemize}

\subsection{Lengthier feedback on trials}~\label{app:lengthier_feedback}
Since the main text summarized feedback on leading signals, for the curious reader, we copied here articulate comments received for each trial.
\begin{itemize}
    \item \textbf{NS}: ``I think this time the arms were not moving? I could not feel any force. I preferred when it did as it indicates the direction without having to look at the feet or wait for the robot to step.''
    \item \textbf{HD}:
    \begin{itemize}
        \item ``Just now I noticed that maybe the robot gives you a heads up for the next movement using the hand on the same side of the movement. THAT WAS COOL. It made things so much easier to follow.''
        \item ``I felt the robot is trying to yeet my arms away :')''
    \end{itemize}
    \item \textbf{HW+HD}: ``In block 1 trials 1-4 I have been dancing by looking only at the robot's feet to tell me which step to take next. In this trial I felt arm tugging. That's awesome. With this, I'll try not to look at the robot's feet and make the steps in the next trials by moving my feet based on the arm tugging initiated by reem-c.''
    \item \textbf{HW+TR}: 
        \begin{itemize}
            \item ``I noticed that the robot was moving its arms/hands slightly before it took each step this time, so I was timing my steps based on that instead of memory (like the second trial) or watching its feet (like the first trial).''
            \item ``The arm tugging is subtle but helpful, because this is the first trial where I danced without looking at reem-c's feet.''
        \end{itemize}
    \item \textbf{HW+HD+TR}: ``The robot was leading with hands too much - hard to tell whether it wants me to move or turn''
    \item \textbf{SC}: ``If it's ``step back step forward etc.'' vs ``1 2 3 4 5 6,'' I prefer ``1 2 3 4 5 6'' like what was done in this trial. It's less confusing in my opinion. I noticed there's no arm tugging this trial (can't tell in block 2 trial 1 because I was distracted by the voice making the commands). As much as I have previously stated how I like the arm tugging, I actually feel ok without the arm tugging. Maybe it's because I'm more familiar with the steps and also interacting with reem-c at this point.''
    \item \textbf{SD}: 
    \begin{itemize}
        \item ``The verbal prompts made me feel more prepared, aware and were engaging. I felt good dancing this time. I guess being guided or spoken to, is a positive effect on the experience.''
        \item ``For the instructions, I prefer 1,2,3 as this is what typically is done for any music related activity that I am familiar with. Stepping instructions are more like when attending a fitness class.''
        \item ``I didn't like how it tells me what to do. I liked the 1,2,3.. approach.''
    \end{itemize}
    \item \textbf{SC + HW}: ``The arm movements were much better. I am not entirely sure this is exactly the same that happened in the other blocks, but this time I realized the arms were moving before the feet had any movement and were leading slightly towards the next direction of movement.''
    \item \textbf{SC + HD:} ``I thought it's realistic that there's arm tugging and verbal cues of ``1 2 3 4 5 6'' when dancing with reem-c. It actually felt like dancing with an instructor (I think... I mean... I never really danced nor have I taken waltz classes, but at least this is what I saw from TV shows and movies).''
    \item \textbf{SC + TR}: 
    \begin{itemize}
        \item ``This one didn't have any hand prompts which was better than the trial with such prominent prompts. I have done it over and over so I guess it's ok, but I would've liked the little hand prompt not with too much force.''
        \item ``The least mental effort in following the robot's moves.''
    \end{itemize}

\end{itemize}

 \begin{table}[bt]
        \centering
        \caption{Summary of answers provided by participants when asked ``What else would you have liked when dancing?''. (\#) reports the number of participants who mentioned a specific aspect when that number is more than one.}
        \begin{tabular}{p{2cm}   p{4cm}}
            \textbf{Topic} & \textbf{Detailed description} \\
            \hline
            
            \rowcolor{gray!20}
            \raggedright Lead work & Better connection (2), head movement (2), indication that the robot starts moving (2), the same but better (2),  lean in the direction of step, clearer weight shifts, smoother motions \\
            
            Dance moves & Different step sequences, turns, or more complex movements (7), closed position, larger steps \\

            \rowcolor{gray!20}
            Audio & Music (6) or ticking beat counter, feedback to teach the dance (5), robot humming the rhythm, counting the steps in the initial trials\\
            
            Interaction & Leading the robot, more interaction \\

        \end{tabular}
        \label{tab:post_suggestions_dancing}
    \end{table}

\begin{table}[bt]
        \centering
        \caption{Summary of answers provided by participants when asked ``How could the interaction have been made better?''. (\#) reports the number of participants who mentioned a specific aspect when that number is more than one.}
        \begin{tabular}{p{2cm}   p{4cm}}
            \textbf{Topic} & \textbf{Detailed description} \\
            \hline
            
            \rowcolor{gray!20}
            Hand motion & Slower/smoother/less sudden (6), more resistance to the human partner (2), use hand motion more as a lead signal \\

            Whole-body motion & Less shaky (3)\\

            \rowcolor{gray!20}
            Audio & Friendlier robot voice, have the robot talk with the partner,  less talking from the robot \\

            \raggedright Tactile interface & Soft pads on the hands and fingers (2), placing the robot hands in a more natural pose (instead of flat hands) \\

            \rowcolor{gray!20}
            Visual features & Eyes like the NAO robot \\

            Duration & Make the trials longer \\

        \end{tabular}
        \label{tab:post_suggestions_interaction}
    \end{table}


\subsection{Post experiment feedback}

After completing all the trials, participants were asked ``What else would you have liked when dancing?'' and ``How could the interaction have been made better?''. Answers offer promising avenues for future work, as detailed in Tables~\ref{tab:post_suggestions_dancing} and \ref{tab:post_suggestions_interaction}, but also show that different individuals enjoy different things (for example, one participant mentioned wanting the robot to talk with them, while another wanted the robot to talk less).

At the very end, participants were asked ``Would you dance again with a robot in the future?'' and were given the choice to answer with ``Yes'', ``No'', or a freeform answer. 19/22 participants simply responded ``Yes'', and the remaining 3 participants elaborated their responses, as summarized in Table~\ref{tab:would_dance_again}.

    \begin{table}[h]
        \centering
        \caption{Answers provided by participants when asked ``Would you dance again with a robot in the future?'' after the experiment. (\#) reports the number of participants who provided the same response.}
        \begin{tabular}{p{6cm}}
            \rowcolor{gray!20}
            Yes (19) \\
            ``for experimental purposes sure, for fun? no way'' (1)\\
            \rowcolor{gray!20}
            ``If they are stable and dynamic enough, why not =)'' (1)\\
            ``Yes if new movements are added.'' (1)\\
        \end{tabular}
        \label{tab:would_dance_again}
    \end{table}

\end{appendices}

\end{document}